# DPD-Cancer: Explainable Graph-Based Deep Learning for Small Molecule Anti-Cancer Activity Prediction

Magnus H. Strømme[1], Alex G. C. de Sá[1,2,*], David B. Ascher[1,2,*]

[1]The Australian Centre for Ecogenomics, School of Chemistry and Molecular Biosciences, The University of Queensland, Brisbane, Queensland, 4072, Australia
[2]Computational Biology and Clinical Informatics, Baker Heart and Diabetes Institute, Melbourne, Victoria, 3004, Australia
[*]To whom correspondence should be addressed to D. B. A. Tel: +61 7 336 53891; Email: d.ascher@uq.edu.au. Correspondence may also be addressed to A. G. C. S. at a.desa@uq.edu.au.

**Abstract**. DPD-Cancer is a graph-attention deep learning framework for predicting small-molecule DPD-Cancer is a graph-attention deep-learning framework for predicting small-molecule anti-cancer activity across the NCI-60 panel, trained and evaluated under a strict chemistry-aware data-partitioning scheme. On the hold-out test set, the classifier achieved an Area Under the Receiver Operating Characteristic Curve (AUROC) of 0.87 (95% CI [0.86, 0.88]) and Area Under the Precision–Recall Curve (AUPRC) of 0.73 (95% CI [0.70, 0.76]); per-cell-line regression models for 73 cell lines produced a median Pearson's Correlation Coefficient (Pearson's R) of 0.64 and median Root Mean Squared Error (RMSE) of 0.67 for pGI50-value prediction. Benchmarks against pdCSM-Cancer, MLASM, and ACLPred under matched data conditions yielded consistently higher Matthews Correlation Coefficient (MCC) scores, an occlusion-based attribution analysis confirmed that model explanations were quantitatively faithful to classifier decisions, and an applicability-domain analysis characterised reliability as a function of chemical distance. To facilitate widespread adoption, DPD-Cancer is available as a free, user-friendly web server for unrestricted use at https://biosig.lab.uq.edu.au/dpd_cancer/.

# Introduction

As of 2026, cancer is the second leading cause of death worldwide—with millions of people affected each year—with roughly one in every six deaths being attributed to cancer (World Health Organization, 2025). Cancer research is not trivial. There has been extensive research into the treatment, prevention, and the biology of cancer, and it remains—today more than ever—an active area of research (Wu et al., 2023; Sullivan, 2023; Truhn et al., 2026; Mondal et al., 2025). Despite extensive efforts worth billions of dollars, there is still a dire need for more effective treatments, with current state-of-the-art methods being constrained by cost, efficiency, availability, and/or ramifications associated with treatments making them infeasible for a large number of cancer patients (Wu et al., 2023; Liu et al., 2024). The integration of artificial intelligence (AI) in drug discovery has accelerated target identification, virtual screening, de novo drug design, and pharmacodynamic and pharmacokinetic drug optimisation (Nigam et al., 2021; Xiong et al., 2021; Weiss et al., 2023; Loeffler et al., 2024). Modern machine learning (ML) and deep learning (DL) models can uncover non-linear patterns and high-order dependencies in biochemical data, which are often infeasible to uncover using conventional computational approaches, enabling more informed prioritisation of potential drug candidates (Atasever, 2024; Muratov et al., 2020). A recent review from Zhang et al. (2025) highlighted AI's growing impact across the oncology pipeline, positioning pharmacodynamic prediction as a core use-case for anticancer discovery.

The majority of these AI-driven approaches leverage the National Cancer Institute's cell-line screening panel (NCI-60) (Shoemaker, 2006), a diverse collection of human tumour cell lines used to measure the growth-inhibitory effects (GI50) of small molecules. The NCI-60 has served as a benchmark for identifying cell-line-specific drug sensitivity, but effectively modelling this data requires navigating the complex interplay between molecular topology and cellular context. While prior work has typically focused on the canonical NCI-60 set, we use an expanded resource of 73 cell lines spanning nine tumour tissue types (National Cancer Institute, 2025), enabling DPD-Cancer to span a broader range of cancer genetic heterogeneity. Several NCI-60 studies have reported strong performance, but comparisons are often clouded by validation choices and analogue leakage that overstate generalisation. To address this, we explicitly investigated how different splitting regimes affect performance, quantified the chemical similarity between training and test data, and developed a graph-attention model trained on jointly curated atomic and molecular features for both pGI50-value regression and binary anti-cancer activity classification.

The evolution of predictive modelling for the NCI-60 has progressed from classical machine-learning ensembles to deep-learning paradigms, but predictive performance benchmarks remain notoriously difficult to compare. Early studies, such as the pharmacogenomic models developed by Cortes-Ciriano et al. (2016), used Self-Organising Maps (SOMs) to group chemistry but lacked explicit similarity-constrained boundaries between training and test subsets. This issue persists in more recent frameworks such as pdCSM-Cancer (Al-Jarf et al., 2021), MLASM (Balaji et al., 2024), and ACLPred (Yadav & Kim, 2025), the latter two of which are not based on NCI-60 panel data. While these platforms achieve high AUROC and correlation scores, their reliance on standard random cross-validation frequently masks "analogue leakage", where models rely on structural memorisation rather than biological generalisation. Even when chemistry-aware partitioning is attempted—such as the 70% Tanimoto threshold employed by Vishwakarma et al. (2024)—the proximity of test molecules to the training boundary leads to over-optimistic performance estimates that drop sharply when the model is challenged with truly novel chemotypes. Compounding the validation problem is the limitation of molecular representation: many state-of-the-art methods, including pdCSM-Cancer and ACLPred, operate on static, hand-crafted 1D and 2D descriptors that fail to capture the topological structure of the molecular graph. While these methods offer "explainability" through post-hoc feature importance, they lack the atom-level transparency required for practical lead optimisation, motivating architectures that learn dynamic representations directly from the molecular graph alongside a more conservative, manifold-based validation scheme.

To address these challenges, we introduce DPD-Cancer, an explainable web-based framework built on a graph-attention transformer architecture for predicting anti-cancer small-molecule activity. DPD-Cancer uses a multi-stage, chemistry-aware data-partitioning strategy that yields a more conservative estimate of model performance on truly novel chemical space. Beyond prediction, the framework incorporates an interpretability layer that maps attention weights back onto the molecular graph, allowing identification of chemical substructures associated with the predicted biological response and bridging the gap between high-performance deep learning and practical medicinal chemistry.

# Materials and Methods

Figure 1 illustrates the end-to-end architecture of the DPD-Cancer platform, which is defined by three parts: data engineering, hybrid featurisation, and prediction. Figure 1A (Data Preparation and Splitting) addresses the critical challenge of 'analogue leakage' by transitioning from raw NCI-DTP molecular data to structurally disjoint subsets. This is achieved through a rigorous pipeline involving Extended-Connectivity Fingerprint (ECFP) generation (Rogers & Hahn, 2010) and pairwise distance computation, followed by non-linear dimensionality reduction via Uniform Manifold Approximation and Projection (UMAP) and Hierarchical Density-Based Spatial Clustering of Applications with Noise (HDBSCAN) clustering methods (Campello et al., 2013; Guo et al, 2025). By assigning training, validation, and test folds based on these distinct chemical clusters provided by UMAP and HDBSCAN approaches, DPD-Cancer methodology ensures that the model is evaluated on truly novel chemotypes rather than near-structural analogues.

In Figure 1B (DPD-Cancer Architecture and Featurisation), the model employs a dual-stream learning strategy to capture both local and global chemical contexts. The local stream utilises Graph Attention Transformers (GAT) (Min et al., 2022) with edge-aware attention heads to learn atom-level representations directly from the molecular graph. Simultaneously, a global stream extracts traditional physicochemical descriptors and pharmacophores (Al-Jarf et al., 2021). These feature sets are integrated through a Feature Fusion module with an Adaptive Gating Mechanism (Xiong et al., 2024), which dynamically prioritises information from local and global scales to form a comprehensive blended embedding.

Finally, Figure 1C (Model Output and Task Prediction) demonstrates the multi-task utility of the framework, which performs simultaneous regression models to provide quantitative, cell-line-specific $pGI_{50}$-values and a classification model to determine binary anti-cancer activity.

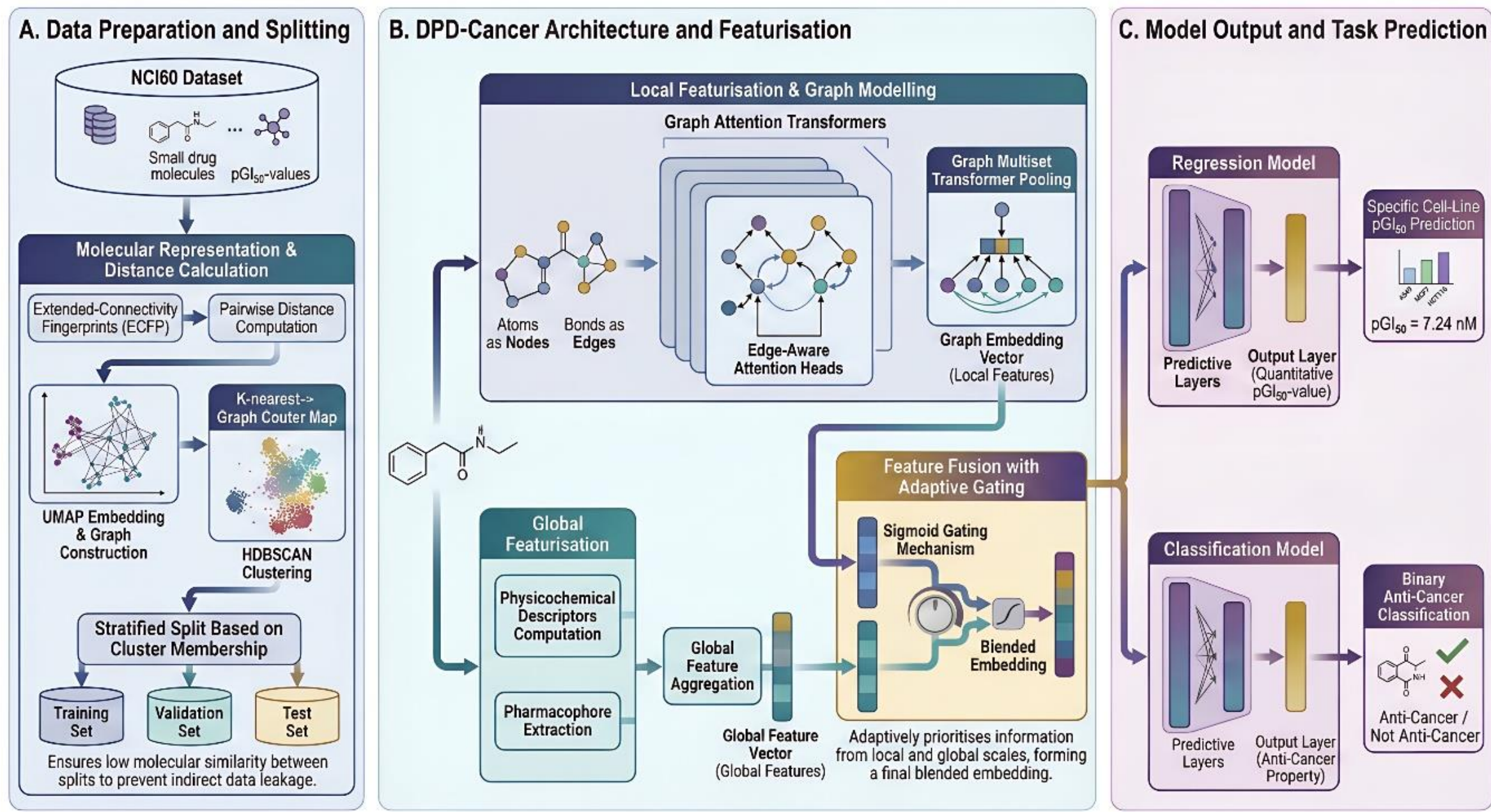

**Figure 1**: **Schematic Overview of the DPD-Cancer framework.** The model utilises the NCI-60 dataset, strictly partitioned into training, validation, and test sets to evaluate generalisability. Molecular structures are processed via graph modelling to capture topological relationships between atoms and bonds. The core framework employs a Graph Attention Transformer. The final embedding is used for dual-task predictions: Binary activity prediction and quantitative $pGI_{50}$-value prediction.

## NCI 60 Data

The "NCI 60 human tumour cell line anticancer drug screen" (NCI-60) dataset developed by the US National Cancer Institute (NCI) in the 1980s, with aims to provide a standardised collection of drug molecule data designed for anti-cancer drug screening. As of April 2025, there are currently 73 distinct cell lines present in the publicly available data set, spanning nine distinct tumour tissue types (breast, CNS, colon, kidney/renal, leukaemia, lung, melanoma, ovary and prostate). The NCI-60 data set provides several potency endpoints– including GI50 (50% cell growth inhibition), TGI (zero net growth), and LC50 (50% cell kill)–computed by interpolation from percent-growth values across a five-dose series. The growth inhibition percentage is the percentage of treated cell growth relative to control growth, normalised for the cell count at the time the drug is added to the NCI-60 assay.

## Data Preprocessing and Partitioning

The $GI_{50}$-value, determined by interpolating the growth inhibition percentage dose–response curve, represents the concentration of the compound needed to inhibit cancer cell growth by 50%. To ensure consistency in data representation, the negative base 10 logarithm of the %$GI_{50}$-values ($pGI_{50}$) are used, such that higher values uniformly reflect increased growth inhibition potency. The molecules' $pGI_{50}$-values were used for anticancer prediction. As tumours from different tissues exhibit distinct molecular features and a unique drug-response profile, the regression analysis was performed on each individual cell line separately.

DPD-Cancer predictive models are developed with two primary aims: continuous $pGI_{50}$-value prediction via regression for potency ranking, and binary classification for rapid anticancer screening. For classification, compounds were labelled "Active" if their average $pGI_{50}$ was $\geq 5$ (effective at $10^{-5}$ molar), and "Inactive" if $<5$. When there were multiple entries for a single canonical Simplified Molecular Input Line Entry System (SMILES) encoding (Weininger, 1988), $pGI_{50}$-value mean imputation was computed and used for downstream analysis. Finally, Butina sphere-exclusion was applied to the data to exclude molecule pairs with a Tanimoto similarity $\geq 0.95$ (Butina, 1999; Tanimoto, 1958). When multiple entries existed for a single canonical Simplified Molecular Input Line Entry System (SMILES) encoding (Weininger, 1988), the arithmetic mean of replicate pGI50 values was used for downstream analysis. Sphere-exclusion clustering was then applied to remove molecule pairs with a Tanimoto similarity $\geq 0.95$ (Butina, 1999; Tanimoto, 1958), ensuring that chemically near-identical molecules not already removed by SMILES deduplication were excluded from the dataset.

The dataset was split into three subsets: training, validation, and a hold-out test set used for final model evaluation. Pairwise distances between Extended-Connectivity Fingerprints (ECFPs) were computed using the Jaccard metric, and these distances were used to construct UMAP's k-nearest-neighbour graph and its low-dimensional embedding (McInnes et al., 2018). HDBSCAN clustering was then applied to the UMAP embedding using cosine distance (Campello et al., 2013), partitioning the dataset such that molecular similarity between the three splits was minimised. This procedure ensures that compounds with similar chemical structures are not distributed across splits, preventing indirect data leakage that would allow the model to be tested on substructures it has already seen during training. This is particularly important for neural-network architectures, which can otherwise learn to track analogue series across folds and inflate performance metrics whenever near-neighbours appear in both training and test sets (Simm et al., 2021; Janela & Bajorath, 2022). For reference, we also report random splitting and Butina-Taylor clustering as baselines, to demonstrate the robustness of the chosen splitting methodology in maximising chemical diversity between splits.

## Feature Engineering

A plethora of features were generated to accommodate the complexities of biochemical data. Prior to feature generation, the molecular SMILES encodings were sanitised by converting each SMILES into its canonical SMILES representation, enforcing standardised structural representation and ensuring consistency in modelling the data. The primary feature set was derived using a Directed Message Passing Neural Network (D-MPNN) featurisation strategy (Ramsundar et al., 2019). This approach is highly versatile for molecular modelling, as it generates a dual representation encompassing both atomic-level features and molecular bond features. Specifically, the model incorporated 133 unique atomic features alongside 14 distinct bond features (Ramsundar et al., 2019) (hereby referred to as local graph features). Complementing this graph-based approach, a broad collection of 200 physicochemical descriptors was extracted to provide a normalised profile of molecular properties (Ramsundar et al., 2019; Landrum, 2026). To capture structural topology, molecular fingerprints were utilised as vector encodings of the molecule's substructures. Extended-Connectivity Fingerprints (ECFP) were generated to represent these structural components, utilising a radius of two, mapped to 2048-bit vectors. To capture the spatial arrangement of functional groups critical for ligand-receptor interactions, pharmacophore modelling was employed (Reutlinger et al., 2013). This involves encoding pairs of potential pharmacophoric points (e.g., hydrogen bond acceptors or donors, and lipophilic groups) separated by varying topological bond distances. The resulting descriptors represent the frequency of these specific pair-types at defined distances, culminating in a fixed-length count vector comprising 210 features. As validated ligand-based representations, these descriptors support scaffold-hopping and encode bioactivity relevant signals essential for predicting potency endpoints like $pGI_{50}$-values.

To standardise model inputs and remove uninformative feature columns, variance-threshold filtration was applied to drop constant features. After filtration, each remaining feature was standardised to zero mean and unit variance to place heterogeneous features on a common scale, preventing variables with larger numeric ranges from dominating the optimisation and improving numerical stability and convergence (Alexandropoulos et al., 2019). Both steps were implemented using scikit-learn (Pedregosa et al., 2011). Full formulas and additional methodological details are provided in the Supplementary Materials.

## Graph modelling

Graph neural networks have proven effective for molecular property and bioactivity prediction by learning directly from molecular graphs via convolution and message passing (Duvenaud et al., 2015; Kearnes et al., 2016; Yang et al., 2019). Given previous success with attention-based graph models in drug discovery, e.g. AttentiveFP for molecular property prediction and GAT-based drug–target affinity prediction models, we we adopted graph attention to capture atom–bond context within small drug molecules (Xiong et al., 2020; Nguyen et al., 2021). The intrinsic assembly of a chemical compounds allows for modelling them as graphs, in which atoms are represented by nodes, and the molecular bonds connecting the atoms, are represented by edges connecting these nodes (Figure 2). Each molecule was modelled as a directed graph, where each node encode the following features: atom type, aromaticity, formal charge, chirality, partial charge, hybridisation, hydrogen-boding (hydrogen bond donor or acceptor), degree, and the number of hydrogen atoms connected by this atom. The edge features are: bond type, conjugation, stereochemical configuration, and whether bonded atoms are within the same ring.

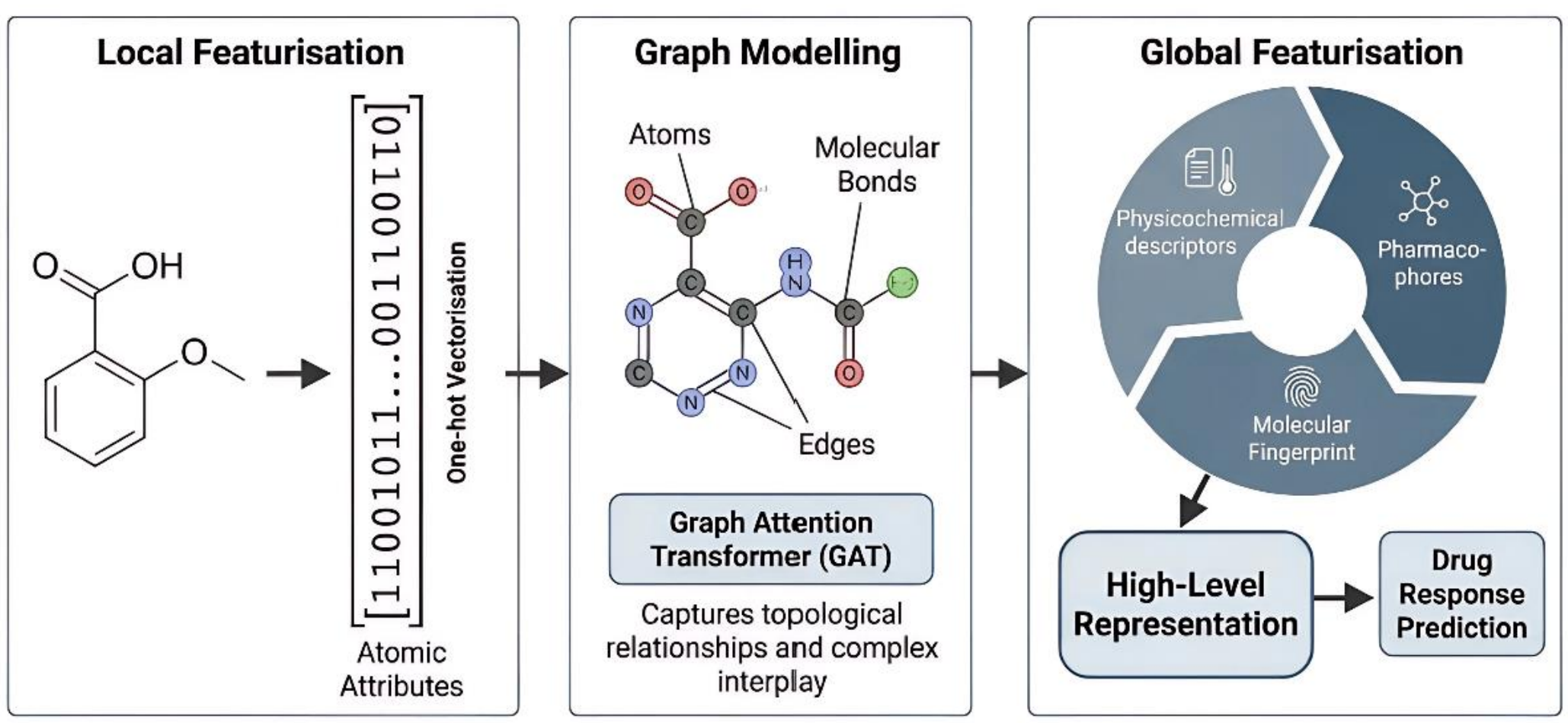

**Figure 2**: **Transformation of Chemical Data**. **Left)** The pipeline begins with local featurisation, where each molecule is converted into atom- and bond-level numerical representations that encode elemental, structural, and chemical properties. **Centre)** These features are organised into a molecular graph, in which atoms and molecular bonds are represented as nodes and edges, allowing the Graph Attention Transformer (GAT) to learn topological relationships and context-dependent interactions through message passing. **Right)** The pooled graph embedding is fused with precomputed global molecular descriptors through a dedicated global-feature branch, yielding an integrated molecular representation for drug response prediction.

## Graph Attention Transformer

Our model is based on an attention–based graph architecture to learn molecular representations from graph-structured chemical features and global physicochemical descriptors for bioactivity prediction. The graph network was implemented with transformer convolutions using the PyTorch Geometric library (Fey et al., 2025; Shi et al., 2021). The Graph Attention Transformer (GAT) merges the permutation-invariance of Graph Neural Networks (GNNs) with the global receptive field of the standard transformer architecture (Min et al., 2022). The hidden state of a node is updated by aggregating features from its neighbourhood, weighted by the learned attention scores. Manual additive skip connections were utilised instead of the built-in gating mechanism within the transformer layers. This approach preserves unhindered gradient flow during backpropagation and maintains architectural symmetry between the graph convolution and the feed-forward network sub-layers (Xiong et al., 2020). Each hidden layers was implemented with a multi-head attention mechanism (Veličković et al., 2018). The multi-headed graph attention mechanism calculates an attention coefficient between connected nodes, determining how much information node $j$ should pass to node $i$ during the message-passing phase. The multi-head attention mechanism is primarily implemented to learn multiple graph feature interactions, i.e., by splitting the n-dimensional features into separate subspaces for each node, the model runs attention mechanisms in parallel over each subspace before averaging the results, theoretically yielding more informative graph embeddings (Shi et al., 2021). The attention heads are averaged rather than concatenated. Graph normalisation was incorporated after each hidden convolutional layer as described by Cai et al. (2021), facilitating faster convergence and improved model generalisation. Dropout—exposing each node to stochastically sampled neighbouring nodes during training—was incorporated into each of the hidden layers to prevent overfitting and improve generalisation of the model (Zhang & Xu, 2024). Global attention-based graph pooling was performed to compress each graph into a single fixed-sized vector, which is then fed to the network. Attention-based pooling was implemented leveraging the Graph Multiset Transformer (GMT) architecture described by Baek et al. (2021).

A gating fusion mechanism was implemented to fuse the two feature scales together to form a blended embedding vector, in line with recent gated graph-attention model designs that use sigmoid gates to regulate information flow across feature scales (Xiong et al., 2024). The gating mechanism involves a sigmoid "gate" which adaptively regulates how the network prioritises information by deciding to keep or discard features from two information sources (local graph features and global graph features). The fusion of local and global graph features allows the network to integrate a chemical compound's atomic features with its molecular-level descriptors, enabling multi-scale structure–activity relationship modelling. Utilising a gating mechanism has several benefits over simple feature concatenation. Beyond simple feature concatenation, the gating mechanism offers two advantages: The learned gate adaptively weights the two feature spaces based on their predictive informativeness for each input, and element-wise mixing remains differentiable, helping the optimiser distinguish between graph and global signals during early training while preventing either input from dominating (Jin et al., 2021; Mungoli, 2023).

Prediction-probability threshold calibration was performed for binary classification to optimise the trade-off between sensitivity and specificity under class imbalance. To further mitigate imbalance, a weighted sampler (Paszke et al., 2019) assigned each sample a sampling probability inversely proportional to its class frequency, such that the minority-class samples were drawn more often during data loading and the model was less biased toward the majority class. The classification model was trained with cross-entropy loss with label smoothing, optimised using Adam with decoupled weight decay (AdamW; Loshchilov & Hutter, 2019). Regression models were trained with the same optimiser to minimise the Huber loss between predicted and observed $pGI_{50}$-values.

## Bootstrap Confidence Interval Analysis

To quantify the stability of reported performance metrics, 95% confidence intervals were estimated via bootstrap resampling of the blind test set. For each metric, 10,000 bootstrap samples were drawn with replacement from the test-set predictions, and the 2.5th and 97.5th percentiles of the sampling distribution were used to define the interval. This procedure was applied to several headline classification metrics (AUROC, AUPRC, Accuracy, Balanced Accuracy, Precision, Recall, MCC, Brier Score, and Expected Calibration Error (ECE)) and to per-cell-line regression metrics (Pearson's $R$, RMSE).

## Model Explainability and Saliency Mapping

To interpret model predictions at the atomic level, we used a perturbation-based attribution method operating on the same graph representation used during inference. For classification models, the explanation target was defined as the raw logit of the positive class. For regression models, the explanation target was the scalar regression output. When several checkpoints were analysed jointly, predictions were averaged across models before computing attribution.

Saliency mapping (Simonyan et al., 2014) encompasses several distinct algorithms, each addressing different limitations in tracing model behaviour. Relying on simple gradients (vanilla backpropagation) often produces noisy highlights due to gradient saturation, where the model's confidence drives gradients to zero. Several more advanced attribution methods address this. Integrated gradients (Sundararajan et al., 2017) calculate the gradients at multiple intervals along a straight path from a baseline input — for a molecular graph, a graph with zeroed node features — to the actual input; integrating these gradients yields attribution scores that accurately reflect each input feature's contribution and eliminate the saturation problem. Gradient-weighted class activation mapping (Selvaraju et al., 2020) uses the gradients flowing into the graph processing layers to produce a coarse localisation map that highlights important regions rather than individual pixels or atoms. Graph-specific attributions (Ying et al., 2019; Pope et al., 2019) partition saliency into node-level attribution, identifying critical atoms, and edge-level attribution, identifying critical bonds.

Because the underlying graph representation includes explicit hydrogens, direct atom-wise attribution can be difficult to interpret visually. We therefore grouped each heavy atom together with all attached explicit hydrogens. To do this, the input SMILES was converted back to an RDKit molecule whose atom count matched the graph used by the model. Each heavy atom defined one group, consisting of the heavy atom itself and any bonded hydrogens. This grouping was used both for attribution and for subsequent faithfulness evaluation. Atomic importance was estimated by heavy-atom-group occlusion. Let $s(G)$ denote the model score for graph $G$, where $s$ is the positive-class logit for classification or the scalar output for regression. For each heavy-atom group $h$, a perturbed graph $G^{\setminus h}$ was constructed by setting the node-feature vectors of all atoms in that group to zero, while leaving the graph topology, edge features, and global molecular features unchanged. The importance of group $h$ was then defined as the score drop

$$\Delta_h = s(G) - s\left(G^{\setminus h}\right) \quad (1)$$

Only positive decreases were retained, i.e.

$$I_h = \max(0, \Delta_h) \quad (2)$$

so that atoms whose removal increased the model score were not treated as positively contributing features. The resulting scores were normalised by the maximum value within the molecule to obtain values on [0, 1]. This procedure yields a local, model-specific attribution: an atom group receives a high score if masking it produces a large decrease in the prediction score. Unlike gradient-based saliency, this method directly measures model sensitivity under explicit perturbation of the learned input representation.

For visualisation, hydrogen-group scores were collapsed back onto the heavy-atom skeleton. The molecule was drawn with RDKit using a black-and-white atom palette, and only the top-ranked fraction of atoms was highlighted in red. By default, the top 20% of heavy atoms were displayed, with circular highlights scaled according to the normalised attribution score. This sparse display was chosen to emphasise the most influential substructures and to avoid the diffuse appearance often observed with dense heatmaps.

To assess whether the explanation reflected model behaviour rather than merely producing a plausible visual pattern, we performed a perturbation-based faithfulness test. Using the same heavy-atom groups as above, the top-$k$ groups identified by the attribution method were simultaneously masked, where $k$ corresponded to a specified fraction of the heavy atoms in the molecule (e.g. 5%, 10%, 20%, and 30%). The resulting drop in model score was compared with the score drop obtained after masking an equal number of randomly selected heavy-atom groups. Random masking was repeated multiple times to estimate a mean and standard deviation. An explanation was considered more faithful when masking the top-ranked groups reduced the model score more strongly than masking random groups. This evaluation does not establish chemical ground truth or biological mechanism; rather, it measures whether the highlighted atoms are genuinely influential for the trained model's prediction. We showcase an example of this method in the Results section.

# Results

## Dataset Assembly and Curation

Drug response data were assembled from the April 2025 release of the NCI-60 screening resource, comprising growth inhibition measurements ($pGI_{50}$) across 73 cell lines spanning nine tumour tissue types. Raw records were filtered to retain only molar-concentration entries with a log high concentration of −4, non-zero GI50 values, and growth-inhibition counts between 0 and 100. All molecular structures were converted to canonical SMILES using RDKit, and entries for which no valid molecule could be parsed were discarded. Where multiple measurements existed for the same canonical structure, replicate $pGI_{50}$-values were aggregated by arithmetic mean to yield a single harmonised response profile per compound. For binary classification, compounds were labelled active when the mean $pGI_{50}$ was at least 5 (corresponding to an effective concentration of $10^{-5}$ M) and inactive otherwise. To reduce near-duplicate redundancy and minimise trivial structural carryover between training and evaluation data, sphere-exclusion pruning was applied using Morgan fingerprints (radius of two, using 2048 bits) with a Tanimoto similarity threshold of 0.95, implemented via RDKit's "Leader_Picker" implementation.

After curation, the final dataset comprised 18,894 unique compounds and 108,072 compound–cell-line pairs. The overall class distribution for the binary classification task was imbalanced, with approximately 23% of compound–cell-line pairs labelled active and 77% inactive. For the regression task, individual cell lines ranged from approximately 600 to over 40,000 compound entries after deduplication, with thirteen cell lines falling below 15,000 entries and receiving a reduced-capacity model variant accordingly (see Supplementary Methods). The curated structure and activity tables, together with the train/validation/blind-test split assignments, are publicly available at https://biosig.lab.uq.edu.au/dpd_cancer/data to support independent replication.

## Chemical-Space Partitioning and Split Integrity

To ensure that predictive performance reflects generalisation to novel chemotypes rather than memorisation of close analogues, we compared three partitioning strategies applied to the curated dataset: random splitting, Butina sphere-exclusion clustering, and HDBSCAN clustering on UMAP projections of ECFP4-derived pairwise distances (Figure 3; Table 1).

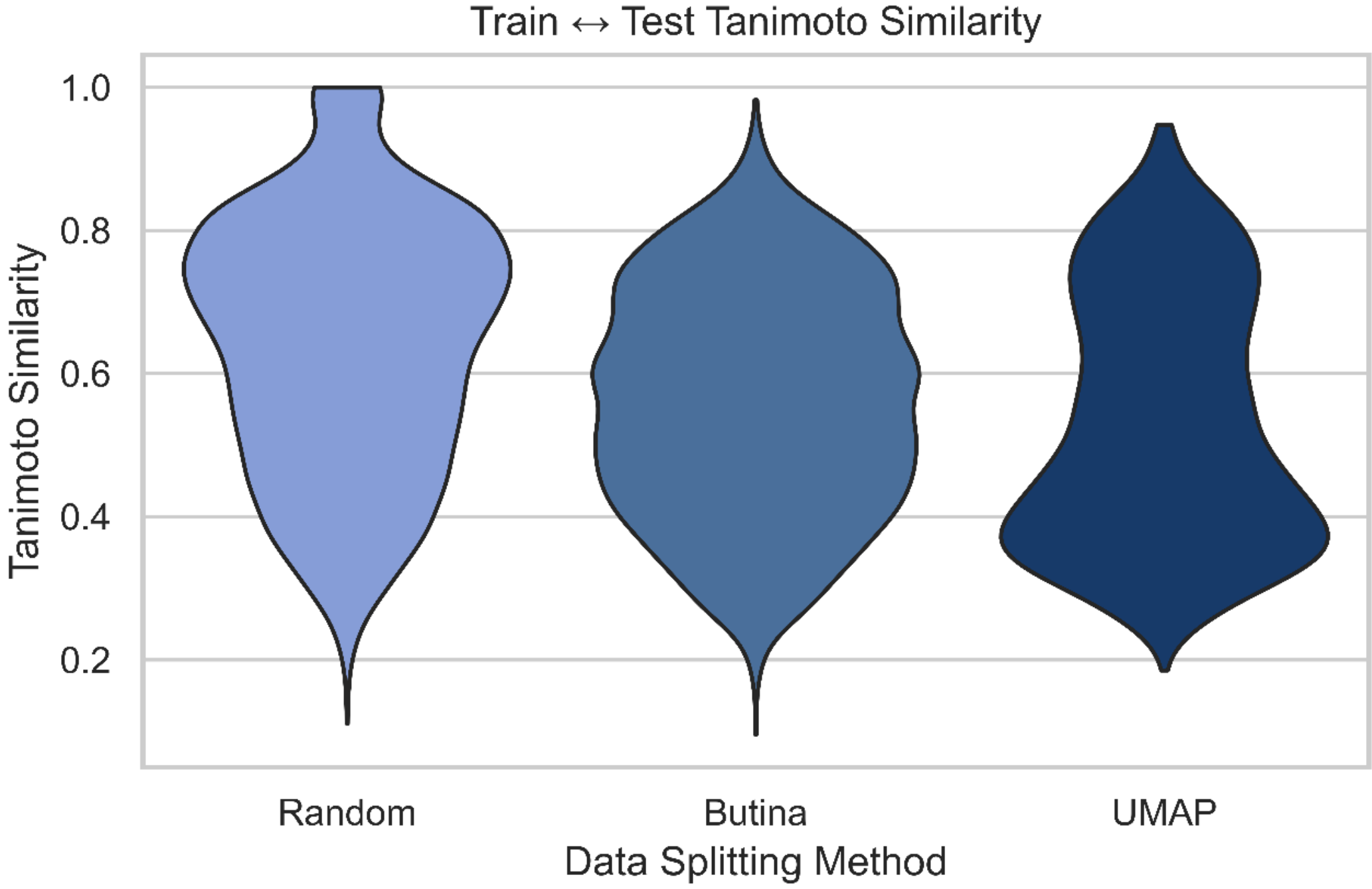

**Figure 3**: **Similarity Distribution of Three Different Data Partitioning Methods**. Random splitting, Butina clustering, and HBDSCAN clustering on UMAP projections. The y-axis displays the Tanimoto similarity distribution between molecule pairs found in the test and training data subsets, i.e., the higher the value, the more similar molecules are across the data used for model training and model evaluation.

Under random splitting, test compounds remained highly similar to the training set (median nearest-neighbour Tanimoto similarity $T_c = 0.66$; mean $= 0.64$), with only 11.3% of the test set exhibiting a maximum Tc < 0.4 to the training data. Butina clustering improved chemical separation, (median $T_c = 0.56$; 17.7% of test set at Tc < 0.4; 63.7% novel scaffolds), but the UMAP/HDBSCAN scheme yielded the strongest chemical separation. For the UMAP-derived test set, the median Tc dropped to 0.50, and crucially, 30.1% of the test compounds shared less than 0.4 Tanimoto similarity with any training molecule. While Butina produced a marginally higher scaffold novelty rate (63.7% vs 61.0% for UMAP), UMAP/HDBSCAN provided a vastly superior reduction in nearest-neighbour similarity, confirming that individual test compounds are substantially more distant from the training set at the level of full molecular structure.

Importantly, analysis of physicochemical properties confirmed that the UMAP split maintains acceptable distributional equivalence between the training and test sets (mean KS p-value = 0.08). This ensures that the increased structural difficulty does not induce significant covariate shift (Supplementary Figure S1). On this basis, all classification and regression results reported below use the UMAP/HDBSCAN partition exclusively. The performance figures presented in subsequent sections should therefore be interpreted as conservative estimates of the model's utility on genuinely unseen chemical matter.

**Table 1**: **Chemical Separation Statistics for Three Data-Partitioning Strategies.** Nearest-neighbour (NN) Tanimoto similarity was computed for each test compound against all training compounds using 2048-bit Morgan fingerprints (radius 2). Novel Test Scaffolds represents the percentage of test molecules whose Bemis–Murcko generic scaffold (Bemis & Murcko, 1996) is entirely absent from the training set. The percentage Technetium (%Tc) <0.4 metric highlights the proportion of the test set that is structurally highly distant from the training data. Mean KS p-value evaluates covariate shift across standard physicochemical properties (values > 0.05 indicate acceptable distributional overlap).

| Partitioning Method | NN Tc (median) | NN Tc (mean) | % Test Set with Tc < 0.4 | Unique Test Scaffolds | Novel Test Scaffolds (%) | Mean KS p-value |
|---|---|---|---|---|---|---|
| Random Split | 0.66 | 0.64 | 11.3% | 3872 | 49.0% | 0.51 |
| Butina Clustering | 0.56 | 0.56 | 17.7% | 3163 | 63.7% | 0.07 |
| UMAP–HDBSCAN | 0.50 | 0.53 | 30.1% | 3186 | 61.0% | 0.08 |

## Anti-Cancer Activity Classification and Benchmarking

To validate the architectural advantages of the proposed GAT framework, DPD-Cancer was assessed using the same benchmark dataset employed by pdCSM-Cancer, MLASM, and ACLPred to isolate the contribution of the model architecture from data-centric variations. For benchmarking against ACLPred, we reconstructed their dataset from the released MLASM data by applying Tanimoto-based redundancy filtering using Morgan 2048-bit fingerprints (radius of two) and with a threshold of 0.85, using global pruning and then performing an unstratified random 80:20 train-test split. As the exact ACLPred filtering procedure was not fully specified, this was used as the closest reproducible ACLPred-style reconstruction. For the reconstructed ACLPred-style benchmark and the MLASM data, limited manual hyperparameter tuning was performed using the training data (of which 15% was used as the model's validation data), to accommodate the vast reduction in dataset size compared to the original dataset used. For simplicity, the hyperparameters used in the training and evaluation of the MLASM and ACLPred datasets, were kept identical.

**Table 2**: **Performance Metrics Classification Model**. Metrics are shown for the hold-out test set reported for each model. DPD-Cancer is also trained and evaluated on the same data set used in the comparisons. Higher score of the comparison pair are marked in bold.

| Method | Evaluation metrics | | | | |
|---|---|---|---|---|---|
| | Accuracy | AUROC | Precision | Recall | MCC |
| DPD-Cancer (updated, strictly partitioned data) | 0.84 | 0.87 | 0.68 | 0.68 | 0.57 |
| pdCSM-Cancer | 0.87 | 0.94 | 0.85 | 0.85 | 0.75 |
| DPD-Cancer (pdCSM-Cancer data) | **0.93** | **0.97** | **0.91** | **0.93** | **0.85** |
| MLASM | 0.79 | 0.88 | 0.81 | **0.74** | 0.58 |
| DPD-Cancer (MLASM data) | **0.81** | **0.89** | **0.89** | 0.70 | **0.63** |
| ACLPred | 0.90 | 0.97 | 0.92 | 0.88 | 0.81 |
| DPD-Cancer (ACLPred data) | **0.93** | **0.98** | **0.93** | **0.89** | **0.86** |

Across all three benchmark comparisons, DPD-Cancer demonstrated consistently superior predictive performance (Table 2). When evaluated on the pdCSM-Cancer dataset, our model achieved an accuracy of 0.93 and an AUROC of 0.97, outperforming the baseline scores of 0.87 and 0.94, respectively. Similar improvements were observed on the ACLPred dataset, where DPD-Cancer reached peak metrics of 0.93 for accuracy and 0.98 for AUROC. Against the MLASM dataset, DPD-Cancer also improved overall accuracy (0.81 vs. 0.79) and AUROC (0.89 vs. 0.88), though MLASM retained a slight advantage in Recall (0.74 vs. 0.70). Crucially, DPD-Cancer achieved higher precision and MCC scores across all comparative datasets. Because MCC is widely regarded as a highly robust metric for bioactivity prediction—particularly concerning class imbalances—the consistent gains in MCC highlight the GAT architecture's enhanced capability to distinguish between active and inactive small molecules without bias. Collectively, these metrics suggest that the transformer convolutions effectively capture complex biochemical and physicochemical features that traditional methods may overlook.

Following the benchmark comparison, we evaluated DPD-Cancer on an updated and strictly partitioned dataset derived from NCI-60 (April 2025 release). This evaluation was designed to test the model's generalisation capabilities under more rigorous screening conditions, where chemical diversity and strict data partitioning impose a greater challenge than the standard benchmark datasets. As expected with a more conservative split, the performance metrics on the updated NCI-60 dataset were lower than those observed on the benchmark data, yielding an accuracy of 0.84 (95% CI [0.83, 0.85]) and an AUROC of 0.87 (95% CI [0.86, 0.88]) (Supplementary Figure S2). The AUPRC of 0.73 (95% CI [0.70, 0.76]) provides a more informative measure of discrimination under the class imbalance present in the dataset. Additional metrics were: precision of 0.68 (95% CI [0.65, 0.71]), recall of 0.68 (95% CI [0.65, 0.70]), MCC of 0.57 (95% CI [0.55, 0.60]), and balanced accuracy of 0.79 (95% CI [0.77, 0.80]) (Supplementary Figure S3). The narrow confidence intervals across all metrics confirm that these estimates are stable and not artefacts of a particular test-set composition. These results tracked closely with the internal validation metrics, confirming that the model generalises well to unseen chemical space. DPD-Cancer's differential performance between the benchmark and the updated NCI-60 dataset illustrates the impact of strict chemical data partitioning: whilst the pdCSM-Cancer, MLASM, and ACLPred datasets permit high theoretical performance, the updated, strictly partitioned results reflect a more realistic approximation of the model's utility in prospective virtual screening.

## Continuous $pGI_{50}$-Value Prediction

Across the NCI panel, the DPD-Cancer regression models achieved consistent and robust performance when tasked with predicting cell line specific drug response from molecular structure. Cell lines with fewer than 600 compound entries were excluded from regression evaluation, retaining 73 cell lines for separate per-cell-line model training. As noted in the Methods, a reduced-capacity architecture was used for the thirteen cell lines with fewer than 15,000 entries. This aspect is also detailed in the Supplementary Methods.

Across all 73 models, performance on the validation and hold out test sets was very similar, indicating good generalisation (Supplementary Table 1). The median Pearson's Correlation Coefficient (Pearson's R) between predicted and observed responses was 0.64 for both the validation and test subsets, with mean values of 0.63 in each case. Validation Pearson's R correlations ranged from 0.42 to 0.80, and test correlations from 0.41 to 0.72. In total, 57 out of 73 models achieved a validation correlation of at least 0.60, and 61 out of 73 models reached at least 0.60 on the test set, with five and seven models respectively exceeding 0.70. Root mean squared error (RMSE) values were tightly clustered, with median RMSEs of 0.67 for both validation and test data and typical values between about 0.5 and 0.8. The lowest test RMSE was observed for CNS line SNB-78 (RMSE=0.46), whilst the highest was obtained for the colon line DLD-1 (RMSE=0.94). Importantly, there was no systematic degradation from validation to test performance. Across all cell lines, the average difference in Pearson's R between test and validation was close to zero, and test correlations were higher than validation correlations for 36 models, lower for 32 models, and identical for five models. This pattern suggests that the models are generalising well to the unseen data subsets.

When aggregated by tissue type (Supplementary Figure S4), DPD-Cancer showed broadly comparable performance across the major cancer indications represented. Mean test set correlations clustered in a narrow band, from 0.59 in prostate lines to approximately 0.65 in kidney, leukaemia, ovarian and melanoma lines, with non-small cell lung, colon and CNS models lying in between. Small cell lung models achieved the highest average validation correlation (0.69) with relatively low validation RMSE (mean of 0.53), and maintained good performance on the test sets (mean Pearson's R=0.61, mean RMSE=0.56). These tissue level summaries indicate that the architecture can capture drug response patterns across diverse biological contexts without a clear loss of accuracy in any particular tissue.

Several individual cell lines stood out as particularly well modelled. The colon cell line HT29 and the leukaemia cell line K-562 achieved the highest test set correlations (both Pearson's R=0.72), with relatively low test RMSEs of 0.61 and 0.69 respectively. High test correlations were also observed for CNS line SF-295 (Pearson's R=0.71), ovarian line SK-OV-3 (Pearson's R=0.71), kidney line SN12C (Pearson's R=0.70) and melanoma line SK-MEL-5 (Pearson's R=0.70). The kidney line SN12K1 exhibited the highest validation correlation overall (Pearson's R=0.80), with test performance remaining in the same general range as the other kidney models (Pearson's R=0.61). In several of these cases, strong correlations coincided with comparatively low RMSE, for example SNB-78 (CNS, Pearson's R=0.70, RMSE=0.46), RXF-631 (kidney, Pearson's R=0.64, RMSE=0.48), SK-OV-3 (ovarian, Pearson's R=0.71, RMSE=0.52) and DMS-273 (small cell lung, Pearson's R=0.62, RMSE=0.53), highlighting cell lines in which the model captures both relative ranking and absolute magnitude of response with high fidelity. Only a small subset of cell lines showed weaker predictive performance. The colon line DLD-1 was the most challenging, with a test set Pearson's R of 0.41 and the highest test RMSE (0.94), suggesting substantial residual variability that is not explained by structure alone. Breast line MDA-MB-468 and non-small cell lung line HOP-18 also showed relatively modest test correlations (Pearson's R is 0.43 and 0.48 respectively), although their RMSE values remained within the overall range observed across the panel. Even in these harder cases the models retained at least a moderate linear association between predictions and experimental measurements, and performance on the test data was broadly in line with that observed on the validation splits.

To enable a direct comparison with an existing method, we also benchmarked DPD-Cancer against pdCSM-Cancer using the original pdCSM-Cancer dataset and official train–test split (Supplementary Table 1). Across the 72 shared cell lines, pdCSM-Cancer obtained a mean Pearson's R of 0.58 on the hold out test sets, compared with 0.56 for DPD-Cancer, with median correlations of 0.59 and 0.56 respectively. Performance between the two approaches was strongly correlated across cell lines (Pearson's R=0.76), indicating that both models capture a similar difficulty landscape. DPD-Cancer outperformed pdCSM-Cancer in fourteen cell lines, presented matched predictive performance in six, and was modestly lower in the remaining fifty-two, with an average absolute difference in Pearson's R of 0.04. Importantly, the GAT architecture delivered clear improvements for several cell lines, including HT29 (colon, 0.72 vs 0.65), SN12C (kidney, 0.70 vs 0.59), M14 (melanoma, 0.69 vs. 0.58), ACHN (kidney, 0.67 vs 0.59) and IGROV1 (ovarian, 0.63 vs 0.57), and achieved Pearson's R $\geq 0.60$ in 18 cell lines compared with 15 for pdCSM-Cancer.

Taken together, these results show that the graph-attention regression framework delivers consistent and robust predictive performance across the 73 NCI-60 cell lines, with closely matched validation and test metrics and no clear bias toward any tissue type. The per-cell-line metrics in Supplementary Table 1 identify both well-modelled and more challenging cellular contexts, illustrating that DPD-Cancer generalises across diverse response profiles while maintaining a consistent level of accuracy. Under the matched evaluation protocol, DPD-Cancer is competitive with pdCSM-Cancer overall and provides a more expressive graph-based representation that yields stronger predictions in specific cellular contexts.

# Interpretability

The attribution map in Figure 4 indicates that the classifier relies primarily on the central nitrogen-rich heteroaromatic core and its connection to the left aryl system, whereas the distal trifluoromethyl-substituted side chains and much of the outer aryl ring contribute less to its prediction.

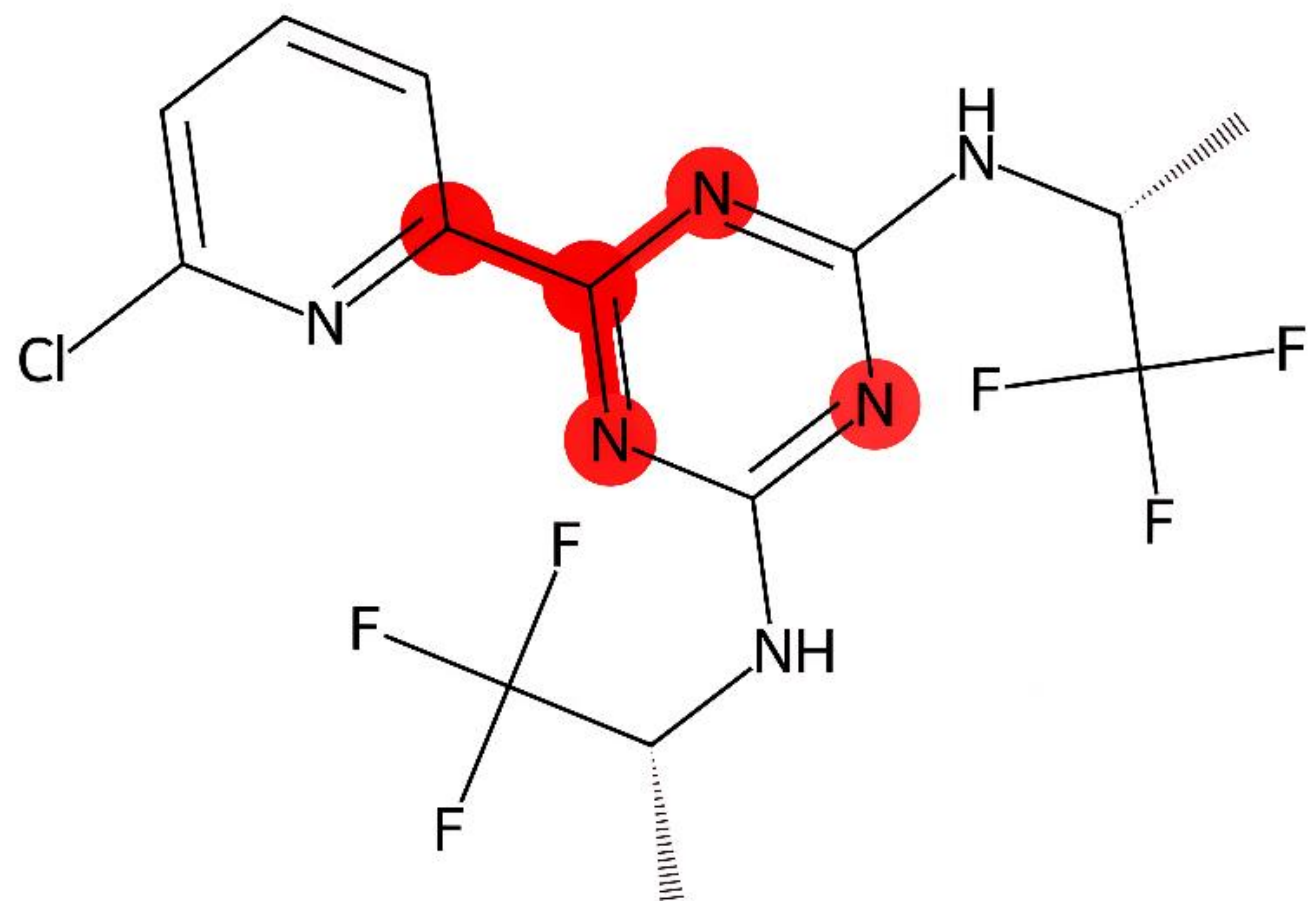

**Figure 4: Occlusion-based Atom-importance Map for the DPD-Cancer Classification Model.** Heavy-atom-group occlusion estimate of the contribution of local substructures to the model's positive-class logit for a recently approved small-molecule cancer drug *vorasidenib* (Voranigo) (approved by the European Medicines Agency (EMA) in July 2025 (Voranigo, European Medicines Agency (EMA), 2025), which was not part of the data set used for training and evaluation of DPD-Cancer. Red circles indicate the top-ranked heavy atoms after grouping each heavy atom with its attached explicit hydrogens and measuring the drop in model score after masking that group.

Faithfulness analysis supported this interpretation: at all masking fractions tested (5%, 10%, 20%, and 30%), masking the top-ranked atoms reduced the model logit more than masking randomly selected atoms of the same size (base logit = 0.83, base probability = 0.78; 5% masking: top-ranked $\Delta = 1.25$ vs. random $\Delta = 0.65 \pm 0.19$; 10%: 1.40 vs. $1.07 \pm 0.16$; 20%: 1.86 vs. $1.39 \pm 0.17$; 30%: 2.07 vs. $1.53 \pm 0.19$). Here, the base logit (0.83) and probability (0.78) are the classifier's original prediction for the intact molecule before any atoms are masked. The masking results then show how much this prediction decreases when selected atoms are removed from the input representation. For example, at 5% masking, removing the top-ranked atoms identified by the interpretation method reduces the logit by 1.25, whereas removing the same number of randomly chosen atoms reduces it on average by only 0.65, with a standard deviation of 0.19. The same pattern held at 10%, 20%, and 30% masking, where masking the highlighted atoms consistently produced a larger drop than random masking.

This indicates that the interpretation method is identifying atoms that are genuinely more important for the model's prediction than an arbitrary subset of atoms. In other words, the highlighted substructure is not simply visually plausible, but measurably more influential for the classifier's decision. These results indicate that the highlighted substructure is meaningfully informative for the classifier's decision rather than merely reflecting a visually plausible pattern. As this explanation is based on systematic perturbation of the input representation, it should be interpreted as a characterisation of model behaviour and feature reliance, rather than as direct evidence of underlying mechanistic causality.

As a proof-of-concept, the attributed atoms in vorasidenib (Figure 4) comprise the 1,3,5-triazine core and its connection to the bridging pyridine–structural elements that overlap with the aminotriazine pharmacophore shared across mutant-IDH1/IDH2 allosteric inhibitors (Mellinghoff et al., 2023). The model does not recover the complete pharmacophore, which in this chemotype additionally involves the hydrogen-bonding NH linkers; nonetheless, the partial overlap demonstrates that occlusion-based attribution can surface chemically meaningful substructures on a held-out approved drug. This is a single-compound illustration rather than systematic mechanistic evidence.

## Applicability Domain and Predictive Reliability

To characterise the domain within which DPD-Cancer predictions are reliable, we examined the relationship between chemical distance from the training set and predictive performance for both classification and regression tasks. For each test compound, we computed the nearest-neighbour Tanimoto similarity ($T_c^{\mathrm{NN}}$) to the training set using 2048-bit Morgan fingerprints (radius 2). Test molecules were then binned by $T_c^{\mathrm{NN}}$ into quartiles, and classification and regression performance metrics were computed within each bin. For classification, Area Under the Receiving Operating Characteristic Curve (AUROC) and Matthews Correlation Coefficient (MCC) were evaluated per bin. For regression, Pearson's Correlation Coefficient (Pearson's R) and Root Mean Squared Error (RMSE) were evaluated per bin across all cell lines pooled.

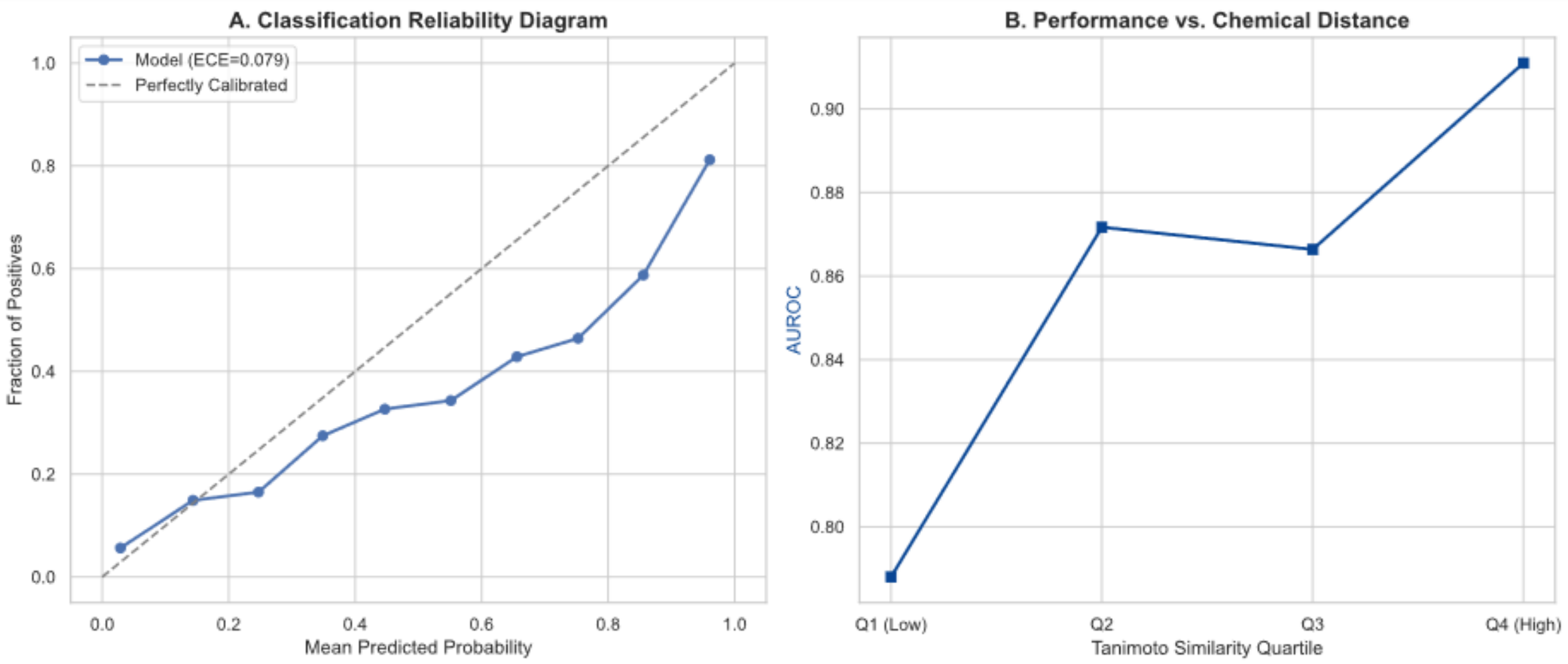

**Figure 5. Applicability Domain and Model Calibration. A)** Reliability diagram for the binary classification model. Predicted probabilities on the blind test set were grouped into ten equal-width bins, and the mean predicted probability was plotted against the observed fraction of active compounds in each bin. The dashed line represents perfect empirical calibration. **B)** Predictive performance as a function of chemical distance. Test compounds were stratified into quartiles based on their nearest-neighbour Tanimoto similarity ($T_c^{\mathrm{NN}}$) to the training set. Classification AUROC (left axis) and pooled regression Pearson's R (right axis) demonstrate how predictive reliability shifts as compounds become structurally further from the training data.

Classification performance was well maintained across the similarity range. Compounds in the highest-similarity quartile ($T_c^{\mathrm{NN}} > Q_3$) achieved an AUROC of 0.91 and MCC of 0.70, while compounds in the lowest-similarity quartile ($T_c^{\mathrm{NN}} < Q_1$) retained an AUROC of 0.79 and MCC of 0.36, representing a decrease of 0.12 in AUROC. This indicates that the classification model generalises to chemically distant compounds with limited degradation, consistent with its intended use for early-stage triage rather than precise activity boundary delineation. To assess the calibration of the classification model, we constructed a reliability diagram by binning predicted probabilities into ten equal-width bins and comparing mean predicted probability against observed positive fraction within each bin (Figure 5). The model exhibited moderate calibration across most of the probability range, with an Expected Calibration Error (ECE) of 0.079 (95% CI [0.070, 0.089]) and Brier score of 0.125 (95% CI [0.118, 0.132]). Mild overconfidence was observed at intermediate predicted probabilities, where the model's predicted positive rate exceeded the observed positive fraction, a pattern consistent with the label-smoothing applied during training and the class imbalance in the dataset. For practical screening workflows, the threshold-calibrated binary output is more reliable than the raw probability magnitude, and we report the calibration diagram (Figure 5A) to enable users to assess confidence independently.

Regression performance showed a more pronounced distance dependence (Figure 5B). For compounds in the highest-similarity quartile, the pooled Pearson's R was 0.82 and RMSE was 0.53, whereas for the lowest-similarity quartile these values shifted to 0.50 and 0.82 respectively. This pattern is expected: quantitative potency prediction depends more heavily on fine-grained structural features that become less transferable as chemical distance increases. The analysis suggests that $pGI_{50}$-value predictions for compounds with $T_c^{\mathrm{NN}} < 0.50$ should be interpreted with greater caution, and we recommend that users of the web server treat regression outputs for compounds far from the training distribution as ranking aids rather than point estimates.

Taken together, these analyses define an empirical applicability domain for DPD-Cancer: classification predictions remain robust across the chemical distance range represented in the blind test set, while regression predictions are most informative for compounds with at least moderate structural similarity to the training data ($T_c^{\mathrm{NN}} \gtrsim 0.50$).

# DPD-Cancer Web Server

DPD-Cancer was developed using the Bootstrap frontend framework and Flask web microframework, hosted on an Apache Ubuntu server. The platform is freely accessible to the research community at https://biosig.lab.uq.edu.au/dpd_cancer. Figure 6 shows the Prediction (Figure 6A), Results (Figure 6B) and Interpretation (Figure 6C).

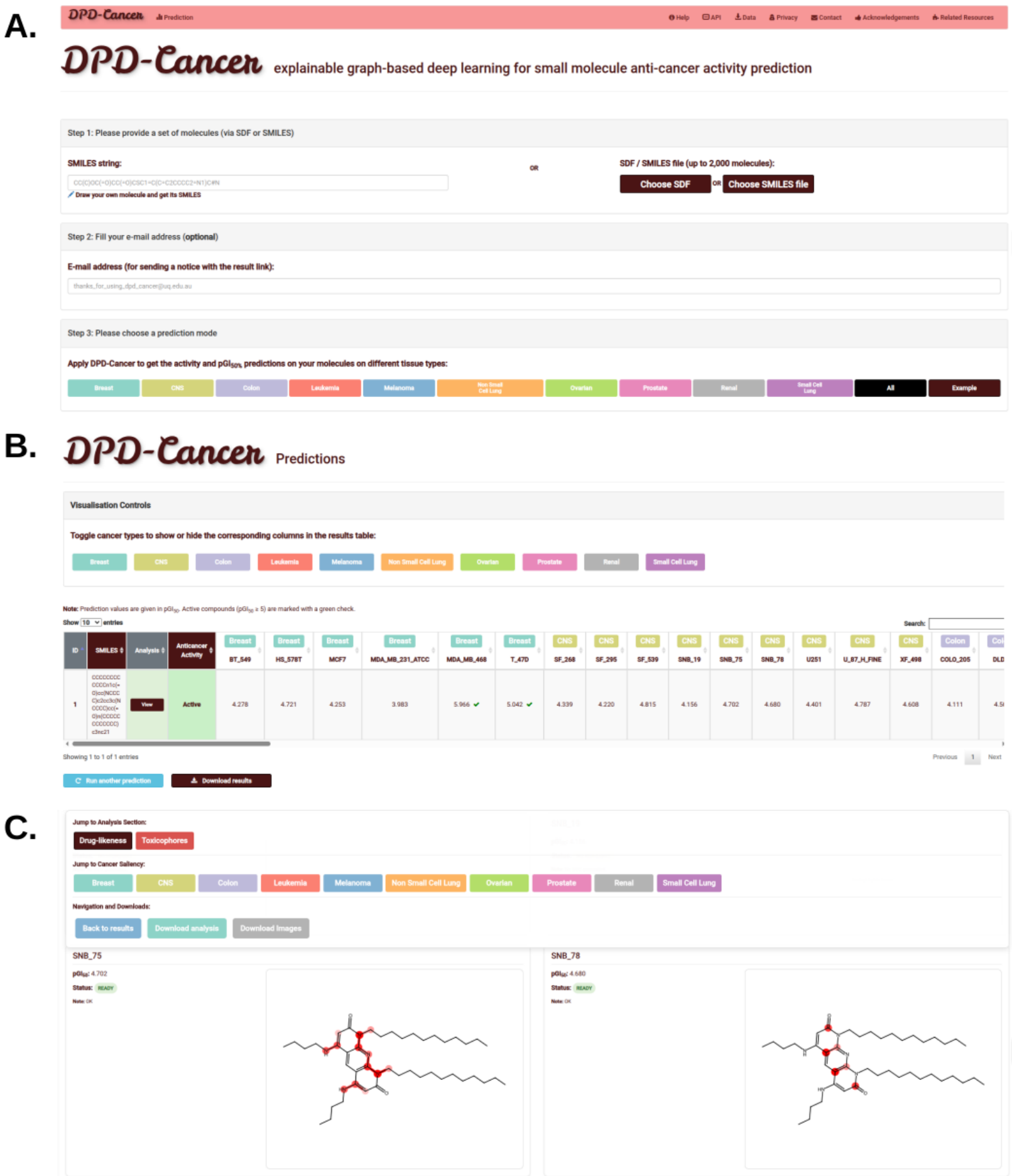

**Figure 6**: **DPD-Cancer web server. A)** presents the prediction page for DPD-Cancer. Users have four input options to provide their molecules: (1) SDF; (2) SMILES file; (3) a single SMILES string; (4) molecular drawing. Users can opt in to include their respective email addresses so the DPD-Cancer web server can send a link with the prediction results. Given the input molecules, users can choose among nine prediction modes, run all of them at once or run an example. As a result, the predictions in terms of anti-cancer activity and $pGI_{50}$ are shown in **B)** in accordance with different cancer types. Users can download the results by clicking on the 'Download results' button. **C)** users can view the analysis and explanation of the prediction by understanding the drug-likeness features, toxicophores and model interpretation (via saliency maps). The whole analysis can be downloaded by clicking on the 'Download analysis' button, or just the saliency maps by clicking on the 'Download images.

**Input**

DPD-Cancer allows users to predict the therapeutic potential and molecular profiles of anticancer candidates. As illustrated in the documentation, the server supports four primary input methods:

- **Single SMILES string:** For rapid analysis of a specific compound.
- **Batch SMILES file:** Supporting high-throughput screening of up to 2,000 molecules.
- **Structural Data File (SDF):** For processing molecular structures in bulk (up to 2,000 molecules).
- **Molecular Drawing Tool:** An interactive canvas where users can sketch structures to automatically generate the corresponding SMILES.

These formats represent the industry standard for cheminformatics research. To assist users, example files and comprehensive documentation are available on the server's, either at **Prediction page** (https://biosig.lab.uq.edu.au/dpd_cancer/prediction) or **Help page** (https://biosig.lab.uq.edu.au/dpd_cancer/help).

**Output**

The DPD-Cancer interface is organised into two distinct modules designed to provide actionable insights:

1. **Prediction Results:** A summary dashboard displaying the core anti-cancer activity scores and property profiles for all submitted molecules.
2. **Analysis Page:** A deep-dive module where users can explore drug-likeness, physicochemical properties, and predictive substructure importance—highlighting which parts of the molecule contribute most to its predicted activity.

**Application Programming Interface (API)**

To support automated drug discovery pipelines and large-scale bioinformatic workflows, DPD-Cancer includes a dedicated **API**.

- **Functionality:** Users can submit molecules via SDF, SMILES files, or strings.
- **Tracking:** Each request is assigned a unique string identifier, allowing users to programmatically retrieve results once processing is complete.
- **Format:** API outputs are delivered in JSON (JavaScript Object Notation) for seamless integration into Python or R environments.

Full documentation, including implementation examples in cURL, is available at the DPD-Cancer API page at https://biosig.lab.uq.edu.au/dpd_cancer/api_docs.

# Discussion

The accurate prediction of drug response remains a formidable challenge in computational biology, primarily due to the complex interplay between molecular structure and cellular context. In this study, we presented DPD-Cancer, a Graph Attention Transformer (GAT) framework designed to capture complex molecular features for both binary classification of compound activity and quantitative prediction of cell-line specific response. Our findings demonstrate that the GAT architecture offers distinct advantages over traditional methods, particularly in extracting meaningful representations from molecular graphs, though the performance depends greatly on the data partitioning.

By benchmarking DPD-Cancer directly against multiple similar methods—pdCSM-Cancer, MLASM, and ACLPred—using their respective identical training and testing sets, the model consistently demonstrated enhanced predictive capacity. Against pdCSM-Cancer, DPD-Cancer achieved higher scores across all classification metrics, with notable gains in Accuracy (0.93 vs. 0.87) and Matthews Correlation Coefficient (0.85 vs. 0.75). Similar performance margins were observed when evaluated against the MLASM and ACLPred baseline data. This consistent outperformance suggests that the attention mechanisms within the DPD-Cancer framework are fundamentally more effective at identifying the specific substructures and atomic characteristics responsible for anti-cancer bioactivity compared to traditional machine learning methodologies. The improvements in precision and MCC across these benchmark datasets highlight the GAT's ability to discern subtle structure-activity relationships, even amidst potential class imbalances. This aligns with recent trends in geometric deep learning, where attention-based models have shown an increased capacity to model long-range dependencies, non-local atomic interactions, and critical functional groups within molecular graphs (Xiong et al., 2020; Zhang et al., 2024). Unlike standard convolution methods that may treat all neighbouring nodes equally, the attention mechanism dynamically weights the importance of specific chemical bonds and atoms, mimicking the targeted biophysical nature of drug-target interactions.

The significant difference between the results on the benchmark dataset and our updated NCI-60 dataset highlights a critical issue in the field: the overestimation of model performance due to data leakage or insufficient chemical diversity in standard benchmarks. Whilst the benchmark metrics were quite high, the performance on the strictly partitioned NCI-60 dataset (Accuracy 0.84, AUROC 0.87) provides a more realistic estimate of the model's utility in *de novo* screening. Importantly, the close agreement between the validation and test metrics (Accuracy 0.83 vs 0.84) confirms that while the task is harder, the model is robust and generalises well to unseen chemical space. This realistic assessment is essential, as models often perform better in theory than they do in actual screening scenarios.

The regression analysis across 73 cell lines revealed that DPD-Cancer is not only a classifier but a capable tool for quantitative response prediction. The strong correlation between validation and test performance across the panel indicates that the model avoids memorising specific training samples—a common pitfall in deep learning models trained on biological datasets with limited sample sizes. Benchmarking against pdCSM-Cancer on the regression task showed that while overall performance was comparable (median Pearson's R of 0.56 vs 0.59), DPD-Cancer provided tangible improvements in specific, biologically distinct contexts. The superior performance in cell lines such as HT29 and SN12C suggests that our graph-based representation may better capture the specific structural motifs relevant to the resistance or sensitivity mechanisms of these tissues. The comparable or slightly lower performance in other lines indicates that for some cellular contexts, the limiting factor may not be the molecular architecture, but rather the availability of high-quality training data or the inherent stochasticity of the biological assays.

The calibration analysis further contextualises model confidence. The classification model achieves strong discrimination (AUROC = 0.87, 95% CI [0.86, 0.88]) but exhibits moderate probability miscalibration (ECE = 0.079, 95% CI [0.07, 0.09]; Brier score = 0.125, 95% CI [0.12, 0.13]), indicating that raw predicted probabilities should not be interpreted as well-calibrated confidence estimates across the full range. For practical screening, we recommend using the threshold-calibrated binary output rather than the raw probability score, particularly in the intermediate probability range. The AUPRC of 0.73 (95% CI [0.70, 0.76]) is the more informative headline metric for the imbalanced classification setting, and the narrow bootstrap confidence intervals across all metrics confirm the stability of these estimates.

Despite these promising results, there are several limitations to consider. First, the drop in Precision and Recall on the strictly partitioned NCI-60 dataset compared to the benchmark suggests that the model still struggles to predict instances where small structural changes lead to large differences in biological activity, particularly when encountered in truly novel chemical scaffolds. Second, while the GAT architecture improves feature extraction, the regression performance varies significantly between cell lines (RMSE 0.46 to 0.94), implying that a "one-size-fits-all" architecture may not be optimal for every tissue type. A further limitation is that the benchmark comparisons against pdCSM-Cancer, MLASM, and ACLPred use data originating from overlapping or closely related NCI-60 screening resources. These comparisons therefore measure architectural advantage under matched evaluation conditions rather than fully independent external validation on compounds from an unrelated screening pipeline. The vorasidenib attribution case (Figure 4) shows plausible behaviour on a recently approved drug not present in the training data, but systematic prospective validation on an external library remains a priority for future work.

The applicability-domain analysis provides a more granular view of these limitations. Classification performance degrades modestly with increasing chemical distance from the training set, confirming that the model learns transferable activity-associated patterns rather than relying exclusively on close-analogue memorisation. Regression, by contrast, shows a steeper distance-dependent decline, consistent with the expectation that quantitative potency estimation requires finer structural resolution than binary activity discrimination. This asymmetry has practical implications: in virtual screening workflows, the classification output is the appropriate primary filter, with regression scores serving as a secondary ranking layer whose reliability decreases for compounds far from the training distribution. Defining this boundary explicitly—rather than presenting a single global performance figure—aligns with the Organisation for Economic Co-operation and Development (OECD) principles for Quantitative Structure–Activity Relationship (QSAR) model reporting ("Guidance Document on the Validation of (Quantitative) Structure-Activity Relationship [(Q)SAR] Models," 2014) and increases the practical utility of the framework for users who need to assess confidence in a given prediction (Muratov et al., 2020). The calibration analysis further supports this interpretation.

# Conclusion

DPD-Cancer represents a robust step forward in molecular graph-based drug response prediction. By validating the model against both standard benchmarks and strictly controlled datasets, we have confirmed that the GAT architecture offers a genuine advantage in encoding chemical structure. While challenges remain in generalising to strictly unseen chemical distributions, the presented framework offers a transparent, reproducible, and highly competitive tool for prioritising anti-cancer candidates in the early stages of drug discovery. Future work will focus on integrating multi-omics data (e.g., gene expression or mutation profiles) directly into the attention mechanism to better contextualise the molecular features. Additionally, exploring pre-training strategies on larger, unlabelled chemical libraries could further enhance the model's ability to generalise to novel regions of chemical space, potentially narrowing the gap between benchmark and real-world performance.

# Data Availability Statement

All data used in this work is publicly available. The curated compound–activity dataset, including canonical SMILES, $pGI_{50}$-values, and binary activity labels, is available at https://biosig.lab.uq.edu.au/dpd_cancer/data. Train, validation, and blind-test split assignments are provided as CSV files with compound identifiers to enable exact reproduction of the evaluation protocol at https://github.com/Magnushst/dpd-cancer. The DPD-Cancer web server is freely accessible without registration at https://biosig.lab.uq.edu.au/dpd_cancer.

# Supporting Information

The Supporting Information is available free of charge at https://github.com/Magnushst/dpd-cancer.

Supplementary Methods: formulas for gated feature fusion (Equations S1–S3), variance filtering and standardisation (Equations S4–S5), AdamW optimiser and Huber loss details, and reduced-capacity architecture specification for low-sample cell lines.

Supplementary Figure S1: Physicochemical Property Coverage (SVG, TIFF)

Supplementary Figure S2: AUROC Curve for the classification model on the blind test set. (SVG, TIFF)

Supplementary Figure S3: Confusion matrix for the classification model on the blind test set. (SVG, TIFF)

Supplementary Figure S4: Regression scatter plots (predicted vs. observed $pGI_{50}$) across nine tumour types with per-tissue Pearson's $R$. (SVG, TIFF)

Supplementary Table S1: Per-cell-line regression performance (Pearson's $R$ and RMSE) on validation and blind-test subsets for all 73 models.

Supplementary Table S2: Per-cell-line regression comparison between DPD-Cancer and pdCSM-Cancer on the pdCSM-Cancer data split.

## Funding

D.B.A. is supported by an NHMRC Investigator (GRNT2041888). Research supported by the NVIDIA Academic Grant Program.

## Conflict of Interest Disclosure

The authors disclose no conflict of interest.

# Supplementary Materials

## Supplementary Methods

### Feature Engineering

The number of GMT query vectors k was adaptively chosen as a fraction of the median node count of the graphs present in the training set (with upper and lower caps) to balance the representational capacity of the model and computational efficiency, using the formula:

$$k = \max(8, \min(64, \lfloor 0.5 \times median(|V(G)|) \rfloor)) \quad (1)$$

where $|V(G)|$ is the node count of graph $G$ and the median is taken over the training set.

$$gate = \sigma(W \cdot G + b), \qquad 0 \leq gate \leq 1 \quad (2)$$

$$X = gate \cdot x + (1 - gate) \cdot G \quad (3)$$

where $W$ is the learned weight matrix from the linear layer of the gating function, $G$ is the global features, and $b$ is the network bias included in the linear layer. The gate allows the model to emphasise which source is more informative for each dimension. The complementary term *(1 – gate)* scales the pooled graph embeddings, improving gradient flow, similar to the gated transform formulation implemented by Srivastava et al. (2015).

AdamW is an adaptive optimiser based on stochastic gradient descent that incorporates non-accumulative weight decay in its gradient updates, applying weight decay directly to the model parameters rather than as L2-regularisation in the loss (Loshchilov & Hutter, 2019). A weight decay of 1e-4 was added to the loss function to lower the effective capacity of the network and improve it generalise to unseen data (Krogh & Hertz, 1991). For the regression task, the loss function was implemented using "SmoothL1Loss" loss function implemented in the PyTorch library (Paszke et al., 2019), which penalises small residuals quadratically (i.e., $|e| < \beta$) and linearly on large ones ($|e| \geq \beta$), making it less sensitive to outliers than mean squared error (MSE) and helps prevent exploding gradients (He et al., 2018).

To standardise model inputs and remove uninformative feature columns, we applied variance threshold filtration to the feature set to filter out constant features values, using the formula:

$$\mathrm{Var}(X_j) = \frac{1}{n}\sum_{i=1}^{n}(x_{ij} - \mu_j)^2, \ where \ \mu_j = \frac{1}{n}\sum_{i=1}^{n} x_{ij} \quad (4)$$

$$\text{keep } X_j \text{ if } \mathrm{Var}(X_j) > 0 \quad \left(\text{drop if } \mathrm{Var}(X_j) = 0\right)$$

We retained $X_j$ if $\mathrm{Var}(X_j) > 0$ (i.e., dropping features with zero variance). This removes constant features that do not contribute to the downstream prediction in the model, such that dimensionality of the data is reduced without losing any crucial information (Alexandropoulos et al., 2019). After filtration, each remaining feature was standardised to zero mean and unit variance in accordance with the following formula:

$$z_{ij} = \frac{x_{ij} - \mu_j}{\sigma_j} \tag{5}$$

$$\text{where } \mu_j = \frac{1}{n}\sum_{i=1}^{n} x_{ij}, \text{ and } \sigma_j = \sqrt{\frac{1}{n}\sum_{i=1}^{n}\left(x_{ij} - \mu_j\right)^2}$$

Such that the now standardised variables satisfies $E[Z_j] = 0$ and $\mathrm{Var}(Z_j) = 1$. Standardisation places heterogeneous features on a mutual scale, preventing variables with larger numeric ranges from dominating estimators and typically improves numerical stability and convergence (Alexandropoulos et al., 2019). Variance filtration and standardisation was implemented using the Scikit-learn Python library (Pedregosa et al., 2011).

To account for varying number of entries within individual cell lines, a smaller model was used for cell lines totalling less than 15000 entries, affecting thirteen cell lines ('MDA-MB-468', 'SNB-78', 'SN12K1', 'XF-498', 'KM20L2', 'HOP-18', 'DLD-1', 'M19-MEL', 'DMS-114', 'DMS-273', 'LXFL-529', 'RXF-631', ‘U-87-H-FINE’).

# Supplementary Figures

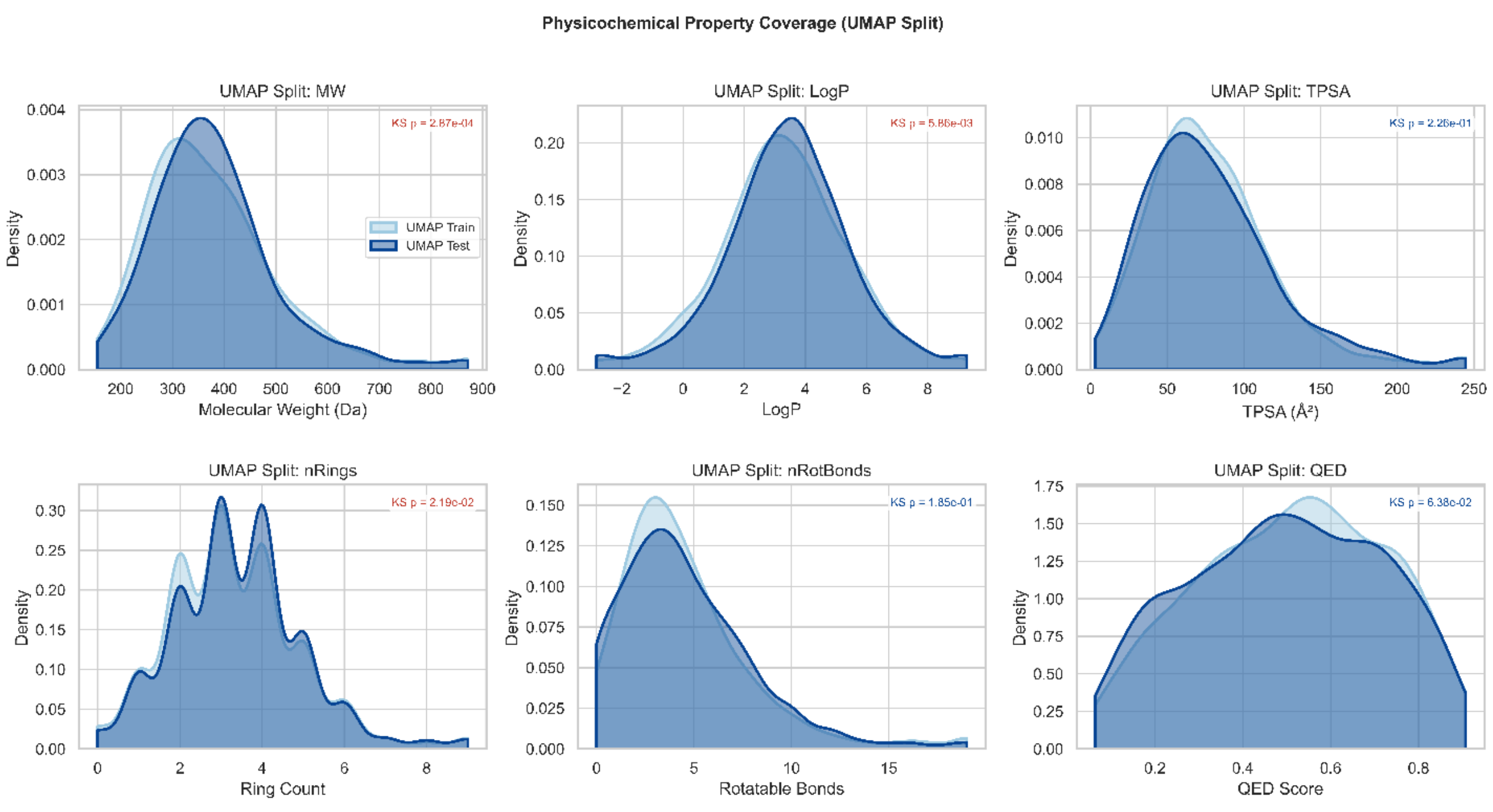

**Supplementary Figure 1. Physicochemical Property Coverage and Distributional Equivalence of the UMAP-HDBSCAN Data Split.** Kernel density estimate (KDE) plots comparing the distributions of the training set (light blue) and test set (dark blue) across six standard molecular descriptors: Molecular Weight (MW), computed partition coefficient (LogP), Topological Polar Surface Area (TPSA), Ring Count (nRings), Number of Rotatable Bonds (nRotBonds), and Quantitative Estimate of Druglikeness (QED Score). The high degree of visual congruence between the sets demonstrates that the UMAP-HDBSCAN partitioning strategy successfully maintains a consistent physicochemical domain across the split. Inset values represent Kolmogorov-Smirnov (KS) test p-values assessing the null hypothesis that the train and test samples are drawn from the same continuous distribution (blue indicates $p \geq 0.05$; red indicates $p < 0.05$). While the high statistical power inherent to large sample sizes yields some significant KS p-values, the macroscopic overlap confirms the absence of severe covariate shift, ensuring that model evaluation reflects generalisation to novel chemotypes rather than artifacts of imbalanced basic properties.

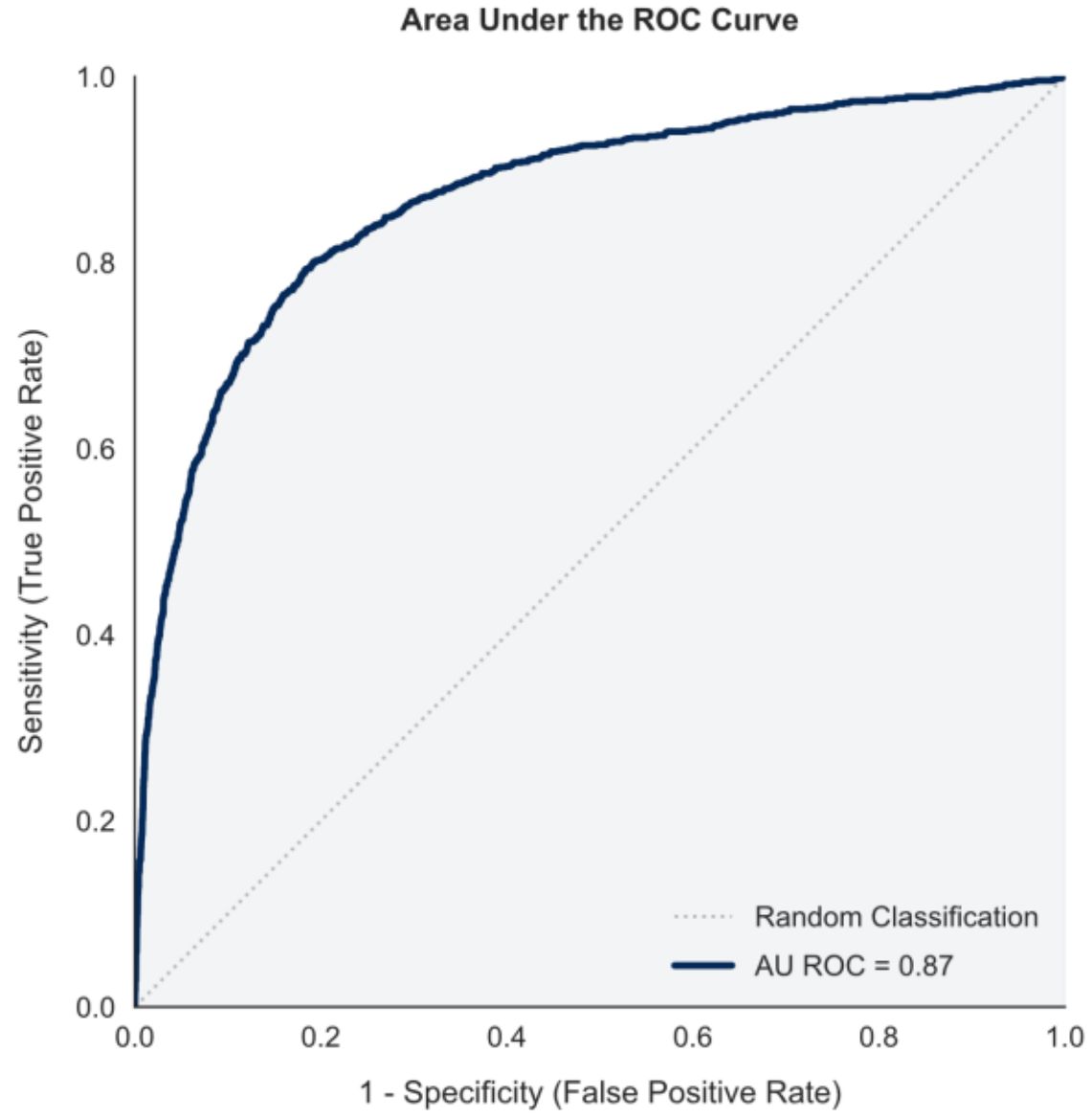

**Supplementary Figure 2. AUROC Curve.** The Receiver Operating Characteristic (ROC) curve plots the model's Sensitivity (True Positive Rate) against 1 - Specificity (False Positive Rate) across various classification thresholds. The solid dark blue line represents the model's performance, achieving an Area Under the ROC Curve (AUC) of 0.87. The dotted grey diagonal line represents the theoretical baseline of a random classifier (AUC = 0.50).

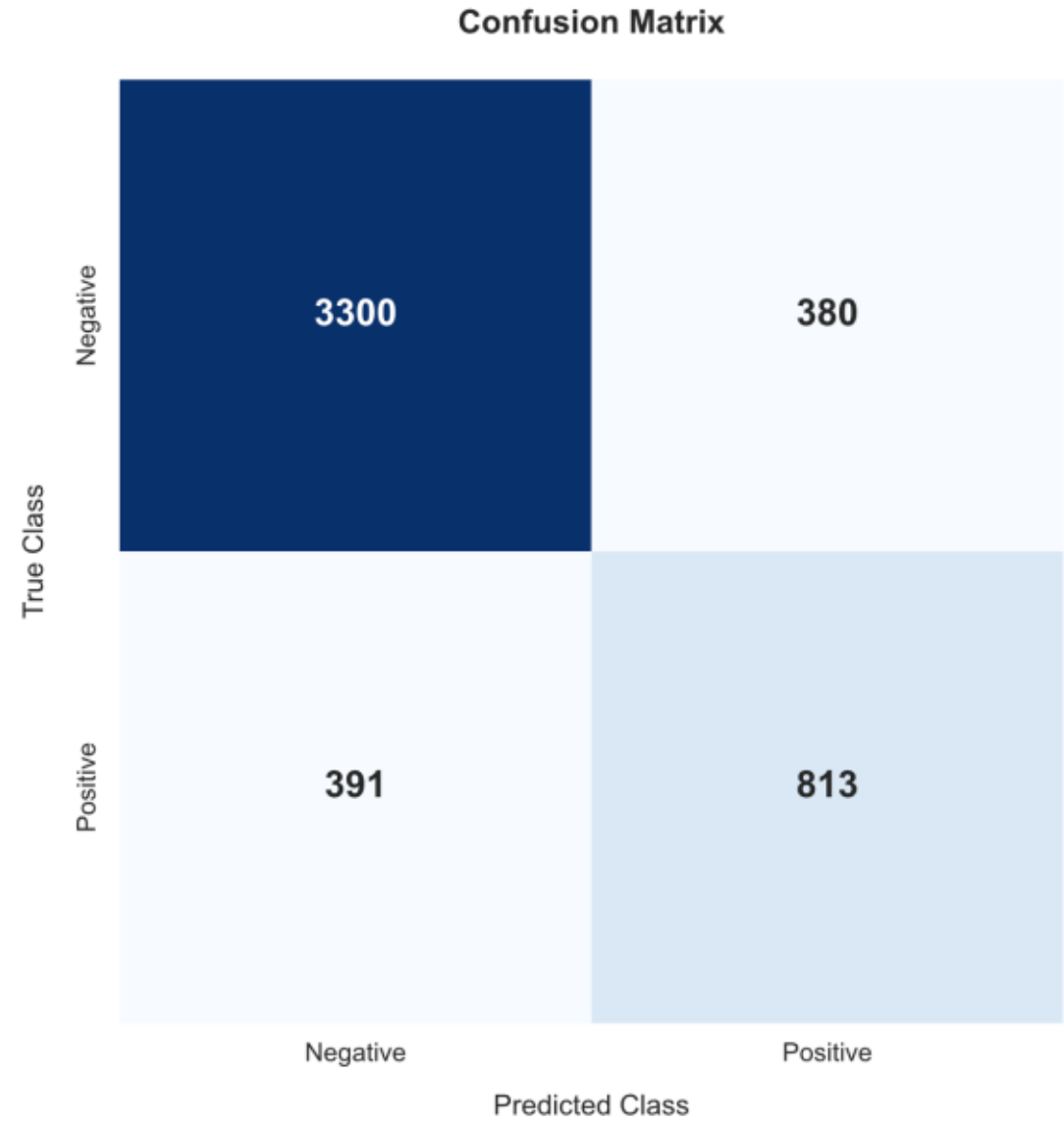

**Supplementary Figure 3. Confusion Matrix.** A cross-tabulation of the model's predicted classes versus the true classes. The matrix displays the absolute counts for true negatives (3300), false positives (380), false negatives (391), and true positives (813), illustrating the model's accuracy in identifying positive and negative instances.

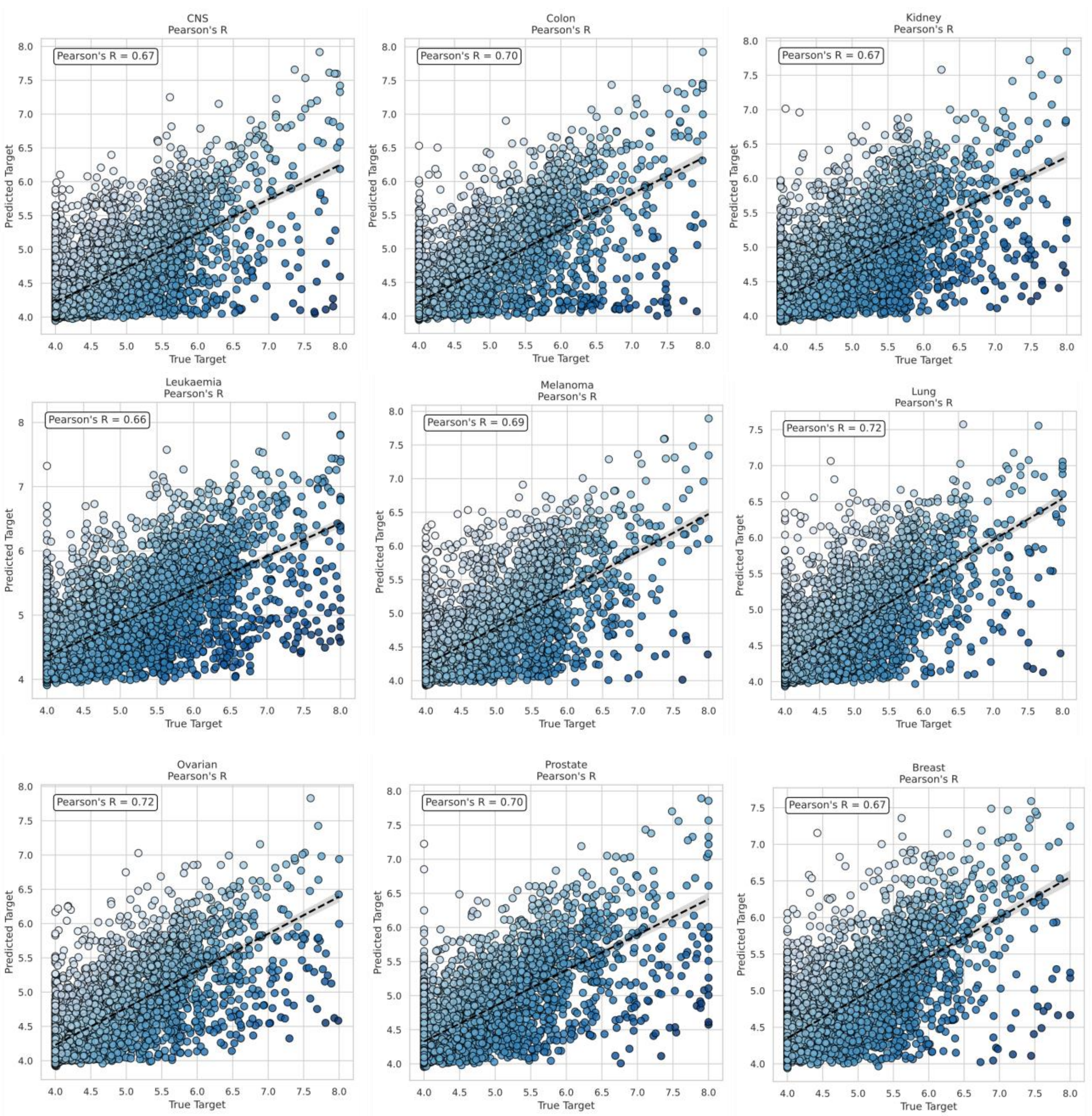

**Supplementary Figure 4**: **Regression analysis of DPD-Cancer predictive performance across nine tumour types.** Scatter plots demonstrating the correlation between predicted and actual anti-cancer activity for small molecules evaluated against nine distinct cancer cell line panels (Breast, CNS, Colon, Leukaemia, Melanoma, Non-Small Cell Lung, Ovarian, Prostate, and Renal). Pearson's Correlation Coefficient (Pearson's R) is provided for each tumour type, highlighting the model's capacity to maintain predictive accuracy across diverse biochemical and cellular environment. The trend line indicates the linear regression fit for the data points.

# Supplementary Tables

**Supplementary Table 1:** Pearson correlation coefficient performance of each modelled cell line. Metrics are shown for the validation subset of the data, as well as the hold-out test set.

| Tissue Type | Cell line | Pearson's R (Validation Data Subset) | Pearson's R (Test Data Subset) | RMSE (Validation Data Subset) | RMSE (Test Data Subset) |
|---|---|---|---|---|---|
| Breast | MCF7 | 0.71 | 0.58 | 0.67 | 0.93 |
| | T-47D | 0.67 | 0.64 | 0.64 | 0.67 |
| | BT-549 | 0.68 | 0.60 | 0.62 | 0.69 |
| | HS-578 T | 0.62 | 0.62 | 0.65 | 0.74 |
| | MDA-MB-231_ATCC | 0.67 | 0.66 | 0.62 | 0.69 |
| | MDA-MB-468 | 0.54 | 0.43 | 0.68 | 0.66 |
| CNS | SNB-19 | 0.55 | 0.66 | 0.73 | 0.59 |
| | SNB-75 | 0.64 | 0.64 | 0.67 | 0.65 |
| | SNB-78 | 0.61 | 0.70 | 0.52 | 0.46 |
| | U251 | 0.66 | 0.67 | 0.69 | 0.64 |
| | XF-498 | 0.48 | 0.52 | 0.66 | 0.65 |
| | SF-268 | 0.61 | 0.66 | 0.61 | 0.62 |
| | SF-295 | 0.67 | 0.71 | 0.63 | 0.61 |
| | SF-539 | 0.68 | 0.64 | 0.67 | 0.67 |
| | U-87-H-FINE | 0.42 | 0.56 | 0.97 | 0.77 |
| Colon | COLO-205 | 0.60 | 0.65 | 0.70 | 0.77 |
| | DLD-1 | 0.53 | 0.41 | 0.61 | 0.94 |
| | HCC-2998 | 0.66 | 0.61 | 0.57 | 0.71 |
| | HCT-116 | 0.65 | 0.65 | 0.76 | 0.69 |
| | HCT-15 | 0.63 | 0.68 | 0.69 | 0.63 |
| | HT29 | 0.71 | 0.72 | 0.67 | 0.61 |
| | KM12 | 0.64 | 0.62 | 0.73 | 0.67 |
| | KM20L2 | 0.59 | 0.58 | 0.61 | 0.54 |
| | SW-620 | 0.68 | 0.69 | 0.65 | 0.70 |
| Leukaemia | CCRF-CEM | 0.63 | 0.65 | 0.80 | 0.74 |
| | HL-60 TB | 0.57 | 0.60 | 0.83 | 0.87 |
| | K-562 | 0.71 | 0.72 | 0.66 | 0.69 |
| | MOLT-4 | 0.60 | 0.64 | 0.79 | 0.70 |
| | SR | 0.65 | 0.63 | 0.76 | 0.90 |
| | RPMI-8226 | 0.69 | 0.67 | 0.62 | 0.72 |
| Non-Small Cell Lung | NCI-H23 | 0.69 | 0.68 | 0.63 | 0.62 |
| | NCI-H226 | 0.63 | 0.63 | 0.65 | 0.61 |
| | A549_ATCC | 0.62 | 0.59 | 0.69 | 0.70 |
| | EKVX | 0.61 | 0.68 | 0.60 | 0.54 |
| | HOP-18 | 0.50 | 0.48 | 0.69 | 0.69 |
| | HOP-62 | 0.65 | 0.66 | 0.64 | 0.64 |
| | HOP-92 | 0.66 | 0.64 | 0.75 | 0.68 |

| | LXFL-529 | 0.64 | 0.59 | 0.50 | 0.60 |
|---|---|---|---|---|---|
| | NCI-H322M | 0.63 | 0.64 | 0.67 | 0.58 |
| | NCI-H460 | 0.68 | 0.69 | 0.73 | 0.64 |
| | NCI-H522 | 0.66 | 0.65 | 0.67 | 0.75 |
| Ovarian | IGROV1 | 0.67 | 0.63 | 0.64 | 0.74 |
| | NCI_ADR-RES | 0.52 | 0.63 | 0.77 | 0.70 |
| | OVCAR-3 | 0.68 | 0.65 | 0.66 | 0.67 |
| | OVCAR-4 | 0.64 | 0.68 | 0.68 | 0.57 |
| | OVCAR-5 | 0.68 | 0.59 | 0.54 | 0.62 |
| | OVCAR-8 | 0.64 | 0.66 | 0.68 | 0.64 |
| | SK-OV-3 | 0.62 | 0.71 | 0.60 | 0.52 |
| Prostate | DU-145 | 0.58 | 0.59 | 0.77 | 0.74 |
| | PC-3 | 0.69 | 0.60 | 0.65 | 0.75 |
| Kidney | 786-0 | 0.64 | 0.65 | 0.71 | 0.65 |
| | A498 | 0.59 | 0.65 | 0.76 | 0.78 |
| | ACHN | 0.71 | 0.67 | 0.64 | 0.61 |
| | CAKI-1 | 0.65 | 0.61 | 0.68 | 0.70 |
| | RXF-393 | 0.57 | 0.68 | 0.84 | 0.65 |
| | RXF-631 | 0.56 | 0.64 | 0.51 | 0.48 |
| | SN12C | 0.67 | 0.70 | 0.60 | 0.58 |
| | SN12K1 | 0.80 | 0.61 | 0.71 | 0.80 |
| | TK-10 | 0.66 | 0.61 | 0.59 | 0.67 |
| | UO-31 | 0.64 | 0.69 | 0.68 | 0.64 |
| Small Cell Lung | DMS-114 | 0.69 | 0.60 | 0.57 | 0.59 |
| | DMS-273 | 0.68 | 0.62 | 0.49 | 0.53 |
| Melanoma | LOX-IMVI | 0.68 | 0.59 | 0.63 | 0.83 |
| | M14 | 0.65 | 0.69 | 0.74 | 0.64 |
| | M19-MEL | 0.52 | 0.63 | 0.71 | 0.48 |
| | MALME-3M | 0.68 | 0.62 | 0.60 | 0.72 |
| | MDA-N | 0.58 | 0.63 | 0.56 | 0.68 |
| | SK-MEL-2 | 0.67 | 0.65 | 0.60 | 0.62 |
| | SK-MEL-5 | 0.65 | 0.70 | 0.72 | 0.62 |
| | SK-MEL-28 | 0.66 | 0.67 | 0.63 | 0.54 |
| | UACC-62 | 0.64 | 0.63 | 0.68 | 0.72 |
| | UACC-257 | 0.58 | 0.69 | 0.68 | 0.61 |
| | MDA-MB-435 | 0.63 | 0.63 | 0.73 | 0.78 |

**Supplementary Table 2**: Pearson correlation coefficient performance of each modelled cell line. Metrics are shown for the hold-out test data set using the same data and data partitioning split used in pdCSM-Cancer.

| Tissue Type | Cell line | Pearson's R (pdCSM-Cancer) | Pearson's R (DPD-Cancer) |
| --- | --- | --- | --- |
| Breast | MCF7 | 0.59 | 0.58 |
| | T-47D | 0.56 | 0.56 |
| | BT-549 | 0.56 | 0.54 |
| | HS-578 T | 0.54 | 0.49 |
| | MDA-MB-231_ATCC | 0.58 | 0.53 |
| | MDA-MB-468 | 0.49 | 0.30 |
| | SNB-19 | 0.61 | 0.59 |
| CNS | SNB-75 | 0.59 | 0.53 |
| | SNB-78 | 0.52 | 0.42 |
| | U251 | 0.63 | 0.67 |
| | XF-498 | 0.51 | 0.42 |
| | SF-268 | 0.59 | 0.60 |
| | SF-295 | 0.59 | 0.58 |
| | SF-539 | 0.59 | 0.55 |
| Colon | COLO-205 | 0.58 | 0.57 |
| | DLD-1 | 0.59 | 0.46 |
| | HCC-2998 | 0.63 | 0.61 |
| | HCT-116 | 0.61 | 0.61 |
| | HCT-15 | 0.56 | 0.51 |
| | HT29 | 0.65 | 0.72 |
| | KM12 | 0.59 | 0.62 |
| | KM20L2 | 0.53 | 0.58 |
| Leukaemia | SW-620 | 0.59 | 0.49 |
| | SR | 0.62 | 0.63 |
| | CCRF-CEM | 0.59 | 0.59 |
| | HL-60 TB | 0.56 | 0.54 |
| | K-562 | 0.57 | 0.54 |
| | MOLT-4 | 0.59 | 0.56 |
| | RPMI-8226 | 0.59 | 0.60 |
| Non-Small Cell Lung | NCI-H23 | 0.59 | 0.53 |
| | NCI-H226 | 0.54 | 0.52 |
| | A549_ATCC | 0.58 | 0.55 |
| | EKVX | 0.53 | 0.55 |
| | HOP-18 | 0.48 | 0.37 |
| | HOP-62 | 0.59 | 0.55 |
| | HOP-92 | 0.57 | 0.55 |
| | LXFL-529 | 0.59 | 0.57 |
| | NCI-H322M | 0.67 | 0.64 |
| | NCI-H460 | 0.63 | 0.60 |

| | NCI-H522 | 0.57 | 0.53 |
|---|---|---|---|
| Ovarian | IGROV1 | 0.57 | 0.63 |
| | NCI_ADR-RES | 0.56 | 0.55 |
| | OVCAR-3 | 0.58 | 0.56 |
| | OVCAR-4 | 0.62 | 0.58 |
| | OVCAR-5 | 0.54 | 0.51 |
| | OVCAR-8 | 0.59 | 0.51 |
| | SK-OV-3 | 0.57 | 0.55 |
| Prostate | DU-145 | 0.58 | 0.53 |
| | PC-3 | 0.59 | 0.57 |
| Kidney | 786-0 | 0.56 | 0.56 |
| | A498 | 0.63 | 0.65 |
| | ACHN | 0.59 | 0.67 |
| | CAKI-1 | 0.56 | 0.58 |
| | RXF-393 | 0.56 | 0.54 |
| | RXF-631 | 0.63 | 0.63 |
| | SN12C | 0.59 | 0.70 |
| | SN12K1 | 0.65 | 0.61 |
| | TK-10 | 0.61 | 0.55 |
| | UO-31 | 0.55 | 0.49 |
| Small Cell Lung | DMS-114 | 0.58 | 0.54 |
| | DMS-273 | 0.48 | 0.40 |
| Melanoma | LOX-IMVI | 0.59 | 0.58 |
| | M14 | 0.58 | 0.69 |
| | M19-MEL | 0.59 | 0.52 |
| | MALME-3M | 0.59 | 0.54 |
| | MDA-N | 0.61 | 0.59 |
| | SK-MEL-2 | 0.56 | 0.54 |
| | SK-MEL-5 | 0.64 | 0.62 |
| | SK-MEL-28 | 0.59 | 0.59 |
| | UACC-62 | 0.59 | 0.57 |
| | UACC-257 | 0.59 | 0.58 |
| | MDA-MB-435 | 0.59 | 0.57 |